\documentclass[]{article}
\usepackage[]{iclr2026_conference}

\iclrfinalcopy

\usepackage{amsmath}
\usepackage{amsfonts}

\usepackage{physics}

\usepackage{graphicx}
\usepackage{wrapfig}
\usepackage{subcaption}
\usepackage{float}

\usepackage{booktabs}
\usepackage{multirow}
\usepackage[labelfont=bf]{caption}

\usepackage[utf8]{inputenc} 
\usepackage[T1]{fontenc}    
\usepackage{hyperref}       
\usepackage{url}            
\usepackage{booktabs}       
\usepackage{amsfonts}       
\usepackage{nicefrac}       
\usepackage{microtype}      
\usepackage{xcolor}         

\usepackage{algpseudocode}
\usepackage[ruled, vlined, linesnumbered]{algorithm2e}

\usepackage{amsthm}
\usepackage{xcolor}
\usepackage{float}

\usepackage{placeins}

\usepackage{mathrsfs}

\usepackage[noabbrev]{cleveref}

\author{%
  Francesco D'Amico$^{1,2}$\thanks{These authors contributed equally to this work.} ,\quad Dario Bocchi$^{1,2*}$,\quad Matteo Negri$^{1,2}$\thanks{\textit{Current address:} LPTM, CY Cergy Paris Université, 2 avenue A. Chauvin, Pontoise 95302}\\ 
  $^{1}$Physics Department, University of Rome Sapienza, Piazzale Aldo Moro 5, Rome 00185 \\
  $^{2}$CNR - Nanotec, Rome Unit, P.le Aldo Moro 5, 00185 Rome, Italy\\
\texttt{\{francesco.damico,dario.bocchi,matteo.negri\}@uniroma1.it} \\
}

\title{Implicit bias produces neural scaling laws in learning curves, from perceptrons to deep networks}

\begin{document}

\maketitle

\begin{abstract}
Scaling laws in deep learning  --  empirical power-law relationships linking model performance to resource growth  --  have emerged as simple yet striking regularities across architectures, datasets, and tasks. These laws are particularly impactful in guiding the design of state-of-the-art models, since they quantify the benefits of increasing data or model size, and hint at the foundations of interpretability in machine learning. However, most studies focus on asymptotic behavior at the end of training. In this work, we describe a richer picture by analyzing the entire training dynamics:  we identify two novel \textit{dynamical} scaling laws that govern how performance evolves as function of different norm-based complexity measures. Combined, our new laws recover the well-known scaling for test error at convergence. Our findings are consistent across CNNs, ResNets, and Vision Transformers trained on MNIST, CIFAR-10 and CIFAR-100. Furthermore, we provide analytical support using a single-layer perceptron trained with logistic loss, where we derive the new dynamical scaling laws, and we explain them through the implicit bias induced by gradient-based training.
\end{abstract}

\section{Introduction}
Neural scaling laws have emerged as a powerful empirical description of how model performance improves as data and model size grow. The first kind of scaling laws that were identified show that test error (or loss) often follows predictable power-law declines when plotted against increasing training data or model parameters.
For example, deep networks exhibit approximately power-law scaling of error with dataset size and network width or depth, a phenomenon observed across vision and language tasks \citep{Hestness2017, sun2017revisiting, rosenfeld2019constructive}. 
Such results highlight the macroscopic regularities of neural network training, yet they largely summarize only the \textit{end-of-training} behavior.

Since the advent of large language models, neural scaling laws started to include the training time, especially in the form of computational budget spent to train a given model.
A seminal work \citep{Kaplan2020} demonstrated that cross-entropy loss scales as a power law in model size, data size, and compute budget, up to an irreducible error floor. These empirical neural scaling laws, including those for generative modeling beyond language \citep{Henighan2020}, indicate a remarkably smooth improvement of generalization performance as resources increase. 
The main interest of this research line is, given a fixed compute budget, to find optimal way to allocate it between model size and training data such that final performance is maximized \citep{Hoffmann2022}. 
Even though empirical results show clean scaling laws spanning for many decades, in particular for language models, there are cases where there are different regimes with different exponents \citep{broken_neural}.
A review that compares various recent methodologies for measuring neural scaling laws can be found in \cite{survey_scaling_laws}.
Notably, in contradiction to scaling laws, which are scale-free, some capabilities of large language models emerge at a certain scale \citep{emergent}. However, it is debated if such phenomena are intrinsic properties of scale or rather of the metrics used \citep{emergent_abilities}.

A complementary line of research studied the so-called \textit{implicit bias} of gradient-based learning dynamics. Implicit bias refers to the inherent tendencies of optimization algorithms to favor certain types of solutions, even without explicit regularization or constraints. For instance, gradient descent often finds solutions that generalize well in overparameterized models \citep{Neyshabur2014, Zhang2017, arnaboldi_escaping_2024}. 
Theoretical results have shown that for linearly separable classification tasks, gradient descent on exponential or logistic losses converges in direction to the \textit{maximum-margin} classifier 
\citep{Soudry2018}, and analogous bias toward maximizing margins has been proven for deep homogeneous networks such as fully-connected ReLU networks \citep{Lyu2020} as well as certain wide two-layer networks \citep{Chizat2020}.

In this work we join these perspectives together by asking whether the implicit bias of gradient descent might itself induce predictable scaling behavior throughout the training process, in models trained with logistic loss. 
The results are organized as follows.
Section~\ref{sec:results_perceptron} focuses on perceptrons.
We first observe a surprisingly good agreement between dynamical learning curves and analytical predictions from the static models with norms fixed at values corresponding to each training stage. We interpret this agreement as a \textbf{training-time implicit bias}.
Then we use the analytical predictions to highlight \textbf{new dynamical scaling laws}, by plotting \textbf{learning curves as a function of the model’s increasing norm}. 
Finally, we show how the new scaling laws can be used to derive established neural end-of-training scaling laws.
Section~\ref{sec:deep} focuses on deep architectures. By using a generalized notion of norm, we reveal that the \textbf{\textit{same} set of scaling laws is present in deep networks}, consistently across architectures and datasets, robust against alternative choices of norm, training algorithms and regularization (the exponents do depend on those details). 
In section~\ref{sec:discussion} we discuss the limits and potential consequences of these results.

\paragraph*{Related works.}
The perceptron has long been a canonical model in the statistical mechanics of learning. Early work established its storage capacity using replica methods, identifying the critical pattern-to-dimension ratio beyond which classification fails \citep{gardner1987capacity, gardner1988optimal}. Later studies analyzed learning dynamics, including exact convergence times \citep{opper1988learning}, the superior generalization of maximum-margin solutions \citep{opper1990generalization}, and Bayes-optimal learning curves as performance benchmarks \citep{opper1991bayes}. 
Online learning was also investigated, with analyses of sequential updates \citep{biehl1994online}, exact teacher–student dynamics in multilayer and committee machines \citep{saad1995exact,saad1995online}, and Bayesian online approaches \citep{solla1998optimal}.

Our main focus is to highlight the role of the norm growth to describe the learning dynamics, which is a perspective that is absent in the classic works. To do that, we use the solution of logistic regression with fixed norm that was studied in \cite{aubin_generalization_2020}. In our work we present an equivalent calculation that reveals the implicit bias at training time and, as a consequence, the new scaling laws.
The idea that implicit bias can extend to the whole learning trajectory can also be found in \cite{wu_benefits_2025}, restricted to the overparametrized regime. 

Few studies on scaling laws include training time independently of the computational cost. Simple models in controlled settings exhibit a power law in the number of training steps \citep{velikanov2021explicit, bordelon2024dynamical}, favoring the discussion on the trade-off between model scale and training time \citep{boopathy2024unified} that is central to the compute-optimal scalings. 
Particularly relevant is \cite{montanari_dynamical_2025}, where the authors connect the different dynamical regimes of a committee machine to its norm, suggesting that the same ideas that we present in our work can apply even outside the setting of logistic loss.
In fact, in the case of regression with square loss, gradient descent is biased toward minimal $\ell_2$-norm solutions when there are many interpolating solutions \citep{Gunasekar2017}.



\begin{figure}[t]
  \centering
  \begin{subfigure}[b]{0.48\linewidth}
    \centering
    \includegraphics[width=\linewidth]{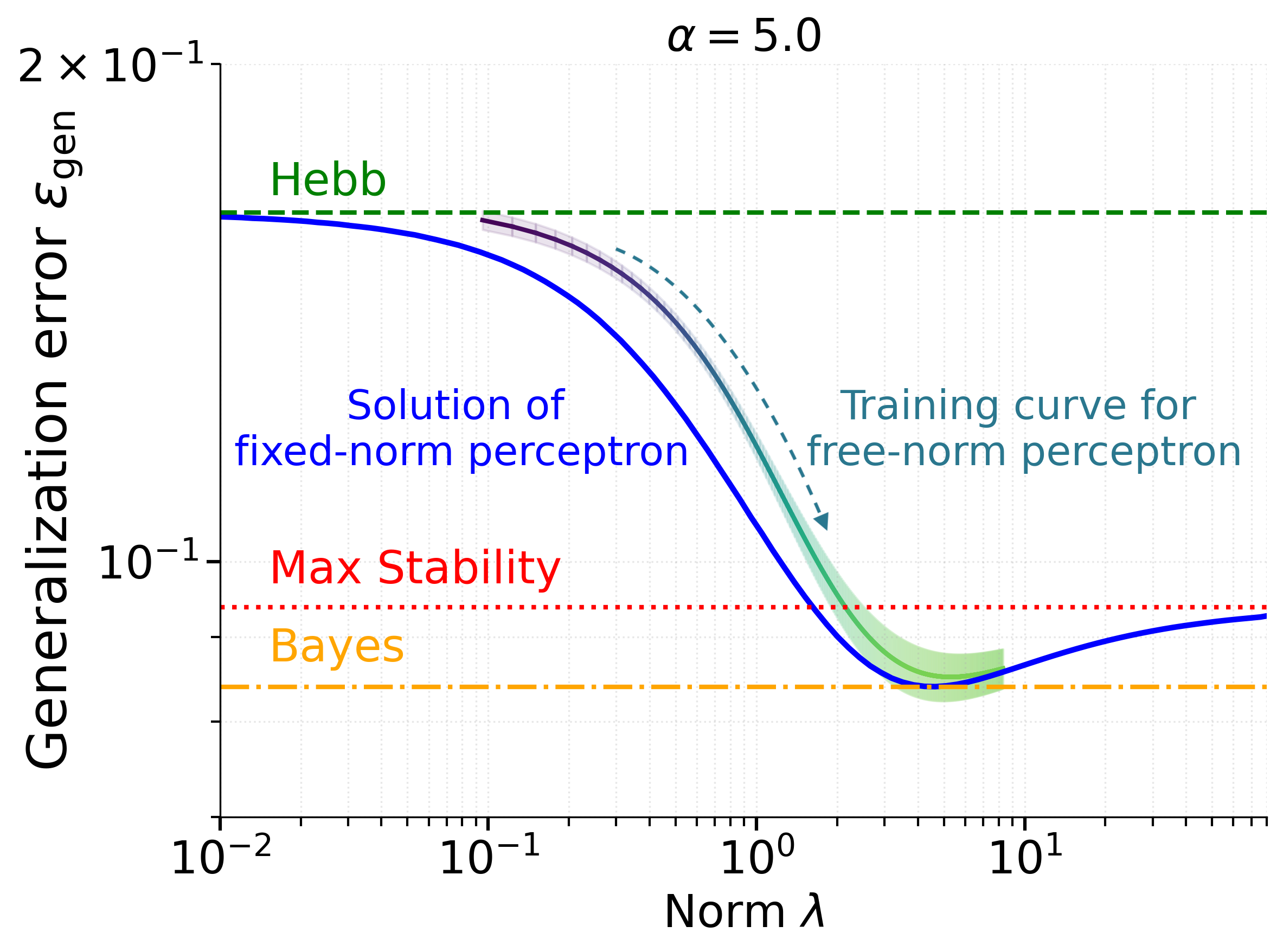}
    \label{fig:boundedVsUnbounded_a}
  \end{subfigure}
  \hfill
  \begin{subfigure}[b]{0.48\linewidth}
    \centering
    \includegraphics[width=\linewidth]{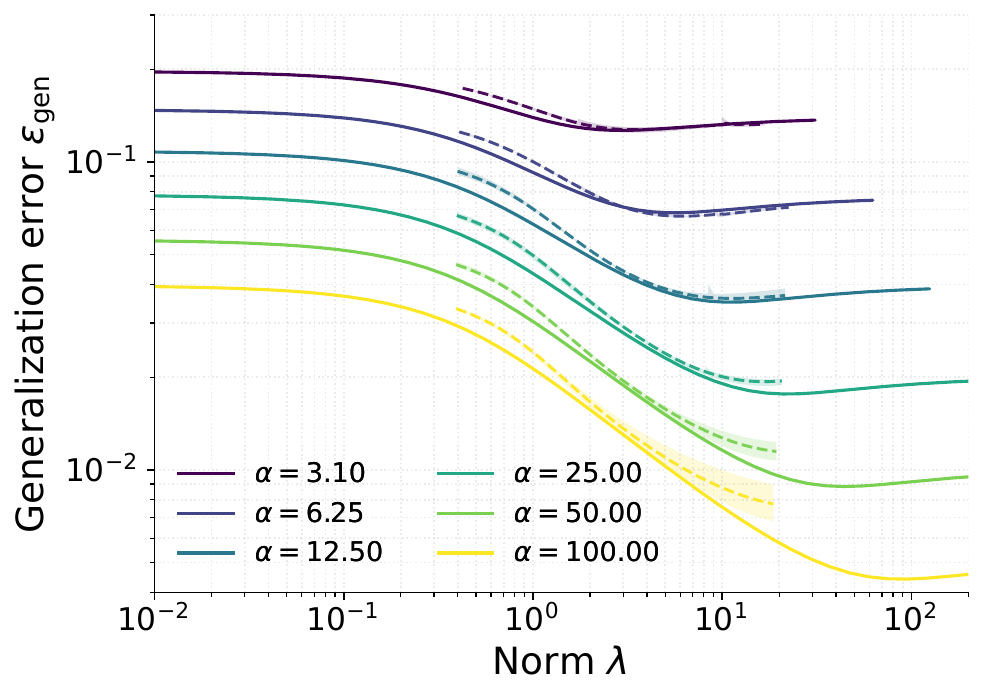}
    \label{fig:boundedVsUnbounded_b}
  \end{subfigure}
  \caption{\textbf{The learning curve of a perceptron with free norm resembles that of fixed-norm problems, which interpolate between known learning rules.} 
  \textit{Left panel:} We plot the generalization error of the minimizers of the cross-entropy loss in a teacher–student setup at a fixed ratio $\alpha=5$ of number of data over size of the system. 
  The blue curve represents the analytical result obtained under a fixed-norm constraint (with $\lambda$ as the hyperparameter of the loss), 
  while the multicolored curve—where color varies with training time—represents the result of numerical training in the free-norm case, where $\lambda$ corresponds to the norm of the weights; the model is trained with $10^6$ steps of gradient descent. The horizontal lines indicate the generalization error of classical learning rules. \textit{Right panel:} Same analysis for different values of $\alpha$; solid curves are analytical solutions at fixed norm,  dashed curves are trajectories with free norm.}
  \label{fig:boundedVsUnbounded}
\end{figure}

\section{Scaling laws in learning curves of perceptrons}
\label{sec:results_perceptron}


This section introduces the core intuitions that we will use for deep architectures -- plotting learning curves as function of the model's norm -- in a setting where we have analytical control of the optimization process.

In the case of a perceptron trained on linearly separable data, it is known that the implicit bias of gradient descent drives the weights toward the maximum-stability solution (the direction that maximizes the classification margin) while the norm grows over time \citep{Soudry2018}. In this section, we ask if the implicit bias has a role \textit{at intermediate stages of training}. Using the well-established teacher–student framework \citep{gardner1988optimal}, we show that the model’s behavior throughout training is qualitatively captured by the solution to the problem in which the norm is held fixed \citep{aubin_generalization_2020}. 
This correspondence allows us to relate the evolution of the perceptron’s norm during training to classical perceptron learning rules, offering a picture on how the implicit bias influences learning dynamics.


\paragraph*{Model definition in Teacher-Student scenario.}

To have an analytical prediction of the generalization error, we consider a framework where a \textit{student} perceptron $\pmb{w} \in \mathbb{R}^N$ attempts to learn an unknown \textit{teacher} perceptron $\pmb{w}^* \in \mathbb{R}^N$ from $P = \alpha N$ labeled examples. Each example $\pmb{x}^\mu \in \mathbb{R}^N$ is a random vector with i.i.d. components $x^\mu_i$ sampled from a Rademacher distribution
$
    P(x^\mu_i) = \frac{1}{2} \delta(x^\mu_i - 1) + \frac{1}{2} \delta(x^\mu_i + 1).
$
The corresponding labels are generated by the teacher as $y^\mu = \text{sign}(\pmb{x}^\mu \cdot \pmb{w}^*)$. We assume both $\pmb{w}^*$ and $\pmb{w}$ to lie on the $N$-sphere, i.e., $\|\pmb{w}^*\|^2 = \|\pmb{w}\|^2 = N$. In this setting, the generalization error (or test error), defined as the expected fraction of misclassified examples on new data, can be written as $\epsilon = \frac{1}{\pi} \text{arccos}(R)$, where $R \equiv(\pmb{w} \cdot \pmb{w}^*)/N$ is the normalized overlap between student and teacher. 
The student minimizes a loss function $L(\pmb{w})$. We study the logistic loss, which reads:
\begin{equation}
    L_{\lambda}(\pmb{w}) = -\sum_{\mu=1}^{P} \frac{1}{\lambda}\left(\lambda \Delta^\mu -  \log 2\cosh\left( \lambda \Delta^\mu \right) \right) = \sum_{\mu=1}^{P}V_{\lambda}(\Delta^\mu),
    \label{eq:pseudolikelihood_loss}
\end{equation}
where we defined the \emph{margin} of the $\mu$-th example as
$
    \Delta^\mu \equiv y^{\mu} \left(\frac{\pmb{w} \cdot \pmb{x}^\mu}{\sqrt{N}}\right),
$
and $\lambda$ is a hyperparameter controlling the sharpness of the logistic loss. Note that for the logistic cost $V_{\lambda}(\Delta)$ we chose the expression in Eq.~\eqref{eq:pseudolikelihood_loss} instead of the more common (but equivalent) $\ln(1+e^{-\lambda\Delta})$ because the former is more convenient to discuss the limits in $\lambda$.
For large $N$, the properties of the minimizers of Eq.~\eqref{eq:pseudolikelihood_loss} can be analyzed via the semi-rigorous \emph{replica method} from the statistical mechanics of disordered systems, which outputs the average value of $R$ from the solutions $\pmb{w}$ that minimize $L_{\lambda}$. In Appendix~\ref{sec:replicas} we present a derivation alternative to that in \cite{aubin_generalization_2020}, where we focus the analysis on the role of the growing norm, which allows us to notice the implicit bias at training time. This observation led us to notice that we can use the norm $\lambda(t)$ as a measure of training status at time $t$, which is one of the key contributions of our work. 

\begin{figure}
    \centering
    \includegraphics[width=0.46\linewidth]{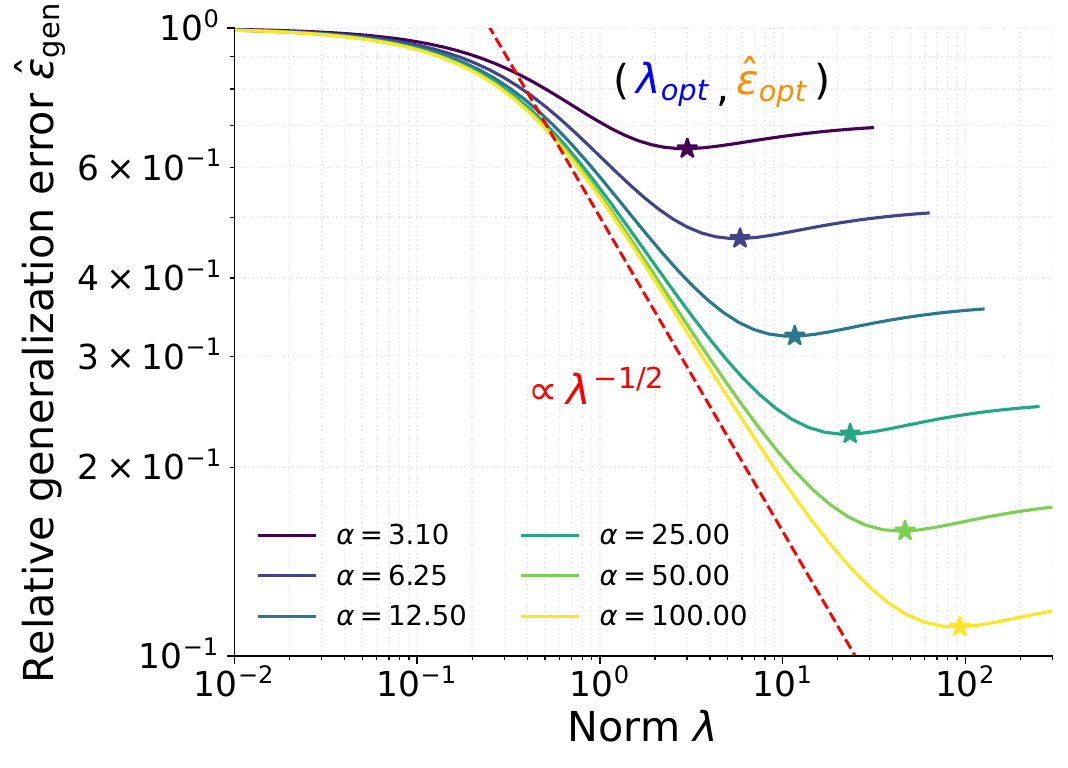}
    \includegraphics[width=0.52\linewidth]{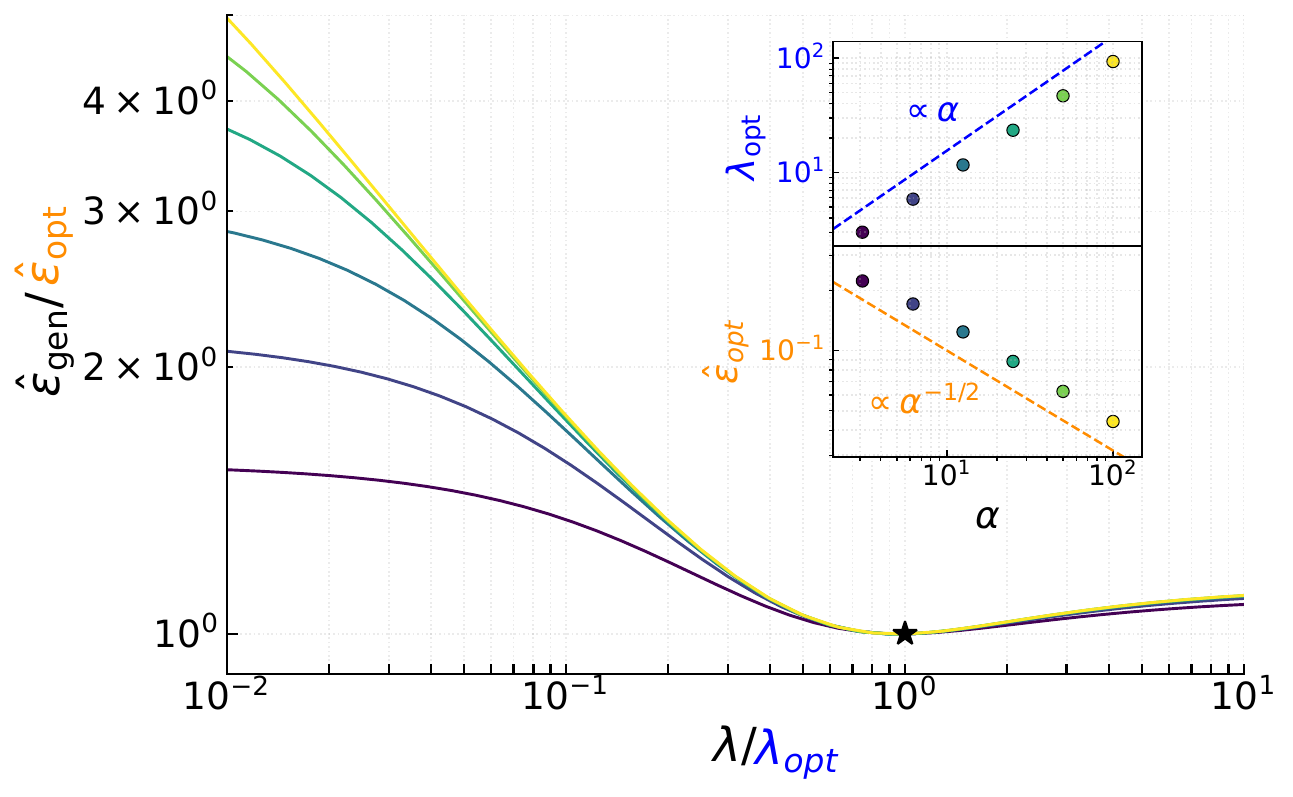}

    \caption{\textbf{Fixed-norm perceptrons exhibit scaling laws in the curves of relative generalization error vs norm.} \textit{Left panel:} we plot the generalization error of the minimizers of the cross-entropy loss in the fixed-norm teacher–student setup of the perceptron, rescaled by the error of the Hebb rule $\epsilon_0$, as a function of the hyperparameter $\lambda$ for different values of $\alpha$. The stars correspond to the optimal points $(\lambda_{\text{opt}}, \hat{\epsilon}_{\text{opt}})$, i.e., the minima of the generalization error for each curve. \textit{Right panel:} we show the same curves after rescaling each one by its corresponding optimal point. The insets display the power-law dependencies of $\lambda_{\text{opt}}$ and $\hat{\epsilon}_{\text{opt}}$ as functions of $\alpha$.}
    \label{fig:twoScalingLaws}
\end{figure}

\paragraph*{$\lambda$-Regimes of the Logistic Loss.}

In Figure~\ref{fig:boundedVsUnbounded}, we show the analytical generalization error as a function of $\lambda$, revealing three regimes:
\begin{enumerate}
    \item \textbf{Small $\lambda$ regime ($\lambda \to 0$):} The second term of Eq.~\eqref{eq:pseudolikelihood_loss} vanishes as $\mathcal{O}(\lambda)$, yielding
    $
        V_{\lambda \to 0}(\Delta) = -\Delta,
    $
    which corresponds (see \cite{Engel2001}) to the Hebbian learning, and defines a baseline generalization error $\epsilon_0$.
    \item \textbf{Intermediate regime and optimal $\lambda$:} At a finite value $\lambda_\mathrm{opt}(\alpha)$, the generalization error is minimum. We find that this optimal $\epsilon_\mathrm{opt}$ matches the generalization error achieved by the Bayes-optimal predictor \citep{opper1991bayes}, suggesting that the logistic loss rule can achieve Bayes-optimality when $\lambda$ is properly tuned. The dependence of $\lambda_\mathrm{opt}(\alpha)$ on $\alpha$ is shown in the top inset of the right panel of Figure~\ref{fig:twoScalingLaws}.
    \item \textbf{Large $\lambda$ regime ($\lambda \to \infty$):} The loss becomes:
    $
        V_{\lambda \to \infty}(\Delta) = -2\Delta\theta(-\Delta),
        \label{eq:gibbs}
    $
    where we defined the step function $\theta(x)=1$ if $x>0$ and $\theta(x)=0$ elsewhere.
    This loss has a degenerate set of minima in $\Delta$ for $\Delta   \geq 0$. In contrast, for any finite value $\lambda$, the minimizer of $V_\lambda(\Delta)$ is unique. For this reason, we cannot apply our method directly to this potential.  To recover the generalization error $\epsilon_\infty$ in the limit $\lambda\to\infty$, one must first solve for finite $\lambda$   and then take the limit $\lambda \to \infty$. We find that this   limiting behavior corresponds to the generalization error of the  maximally stable perceptron
    $
        \pmb{w}_{\text{maxStable}} = \underset{\pmb{w}}{\text{argmax}} \left[\min_{\mu} \Delta^\mu(\pmb{w}) \right]
    $ \citep{gardner1987capacity, opper1990generalization}.
\end{enumerate}

In Figure~\ref{fig:boundedVsUnbounded} we presented curves for $\alpha>1$ because the scaling laws appear more clearly, but the same regimes are present also when $\alpha<1$ (see Fig.\ref{fig:app_overapam} in Appendix \ref{sec:app_overparam}). 

\paragraph*{Norm scaling and interpretation.}

An important observation is that the logistic loss defined in Eq.~\eqref{eq:pseudolikelihood_loss} depends only on the product $\lambda \Delta$ (up to an overall multiplicative factor of $\lambda$ that does not affect the location of the minimizers), where $\Delta$ is linear in the norm of the perceptron weights $\|\pmb{w}\|$. Rescaling the weight norm is thus equivalent to adjusting $\lambda$, meaning that analyzing a fixed-norm perceptron with varying $\lambda$ is equivalent to studying the minimizers of the loss at fixed $\lambda$ and varying norm. This insight also helps explain the behavior of $\epsilon_\infty$: it is known \citep{Soudry2018,Montanari2024NegativePerceptron} that in the infinite-norm limit, the perceptron converges to the maximally stable solution during training (implicit bias).
Building on this observation, we compare two scenarios: the \textbf{fixed-norm} case, where the norm $\|\pmb{w}\|^2 = N$ is fixed and $\lambda$ is treated as a tunable hyperparameter of the loss (the results in this setting are obtained with the replica method); and the \textbf{free-norm} case, where the parameter in the loss is fixed to 1 (i.e., we use the classical logistic loss), and the norm $\|\pmb{w}(t)\| \equiv \lambda(t)$ is left free to evolve during training (here the perceptron is trained using standard gradient descent optimization techniques, and the results are obtained from numerical simulations).

In Figure~\ref{fig:boundedVsUnbounded}, we compare the generalization curves under these two scenarios. 
We remark that in the fixed-norm case, each point on the curve corresponds to the endpoint of training for a different perceptron (at given $\lambda$), while in the free-norm case, the curve represents the trajectory of a single perceptron during training, with each point corresponding to a different time step as the norm evolves.
We see that the free-norm trajectory is qualitatively well described by the set of fixed-norm optimal solutions, indicating that the fixed-norm static analysis captures the essential features of the learning dynamics.

\paragraph*{Scaling laws in learning curves at training time.}
\label{sec:Scaling_laws_fixed_norm_perceptron}

From the left panel of Fig.~\ref{fig:boundedVsUnbounded}, we observe that for sufficiently large $\alpha$ the curves share the same slope but differ in their starting point -- that is the generalization error $\epsilon_0$ of Hebbian learning (for large $\alpha$, $\epsilon_0 \sim \alpha^{-1/2}$). To highlight the power law scaling in $\lambda$, in the left panel of Fig.~\ref{fig:twoScalingLaws} we plot relative error $\hat{\epsilon}_\mathrm{gen} \equiv \epsilon_\mathrm{gen}/\epsilon_0$ as a function of $\lambda$. We observe that for sufficiently large values of $\alpha$, the learning curves of the relative error split into two distinct regimes, which behave differently as we vary $\alpha$.
\begin{enumerate}
    \item \textbf{An early power-law regime, independent of $\alpha$}. The initial part of each learning curves follows the same shape for any $\alpha$, up to a value $\lambda_\mathrm{elbow}(\alpha)$ where it saturates. The  curves collapse for $\lambda<\lambda_\text{elbow}(\alpha)$ on the power law 
    \begin{equation}
        \hat{\epsilon}_\mathrm{gen} = k_1 \lambda^{-\gamma_1} + q_1.
        \label{eq:scaling_law_1_error_vs_norm}
    \end{equation}
    Here we introduce the term $q_1$ to be general, but in the  perceptron we have $q_1=0$. Keeping $q_1$ will be useful in the next section on deep networks, where it we will connect to the \emph{irreducible error floor} of realistic settings.

    \item \textbf{A late regime, which depends on $\alpha$}. After $\lambda_\mathrm{elobw}(\alpha)$, the learning curves deviate from the power law and saturate or overfit following a curve whose height depends on $\alpha$.
\end{enumerate}
It is possible to find proper scalings that collapse also the late-phase curves (actually, the whole training curves will collapse). 
First, we need to discuss the scaling law for the point of minimum test error $\lambda_\mathrm{opt}(\alpha)$. 
In the inset of the right panel of Fig.~\ref{fig:twoScalingLaws}, we observe that the curves follow the power law
\begin{equation}
    \lambda_\mathrm{opt} = k_2 \alpha^{\gamma_2} + q_2,
    \label{eq:scaling_law_2_optnorm_vs_data}
\end{equation}
Like $q_1$, the term $q_2$ is not needed in the fixed-norm perceptron, but we introduce it to obtain a more general law applicable to deep networks.
Now we can compute $\hat{\epsilon}_\mathrm{opt}=\hat{\epsilon}(\lambda_\mathrm{opt})$ and rescale the learning curves of the left panel horizontally by $\lambda_\mathrm{opt}(\alpha)$ and vertically by $\hat{\epsilon}_\mathrm{opt}$ (see Fig.~\ref{fig:twoScalingLaws}, right panel). For large values of $\alpha$, the curves collapse onto a single master curve, i.e.
\begin{equation}
\hat{\epsilon}_\mathrm{gen} / \hat{\epsilon}_\mathrm{opt} = \epsilon_\mathrm{gen} / \epsilon_\mathrm{opt} = \Phi(\lambda / \lambda_\mathrm{opt}),
\label{eq:phi}
\end{equation}
for some universal function $\Phi$. Note that it is a common practice when studying neural scaling laws to drop models trained with too-small datasets (see \citet{survey_scaling_laws}), and the fact that our scaling laws appear only for large $\alpha$ provides a natural justification for this practice.
We also stress that these scaling laws are not a general phenomenon with any choice of loss function: in Appendix~\ref{sec:app_MSE}, as a counterexample, we plot learning curves for Mean Square Error (MSE), which do not show scaling laws.

\paragraph*{Connection to end-of-training scaling law.}
\label{sec:connection_to_literature}

It is tempting to combine the two scaling laws in Eq.~\ref{eq:scaling_law_1_error_vs_norm} and \ref{eq:scaling_law_2_optnorm_vs_data} to recover the well know scaling law $\hat{\epsilon}_\mathrm{gen}(\alpha)\sim \alpha^{-\gamma}$ at the end of training \citep{Hestness2017}.
However, Eq. \ref{eq:scaling_law_1_error_vs_norm} is valid only for $\lambda < \lambda_\mathrm{elbow}(\alpha)$, while $\lambda_\mathrm{opt}(\alpha)>\lambda_\mathrm{elbow}(\alpha)$. Therefore, substituting Eq.~\ref{eq:scaling_law_2_optnorm_vs_data} into Eq.~\ref{eq:scaling_law_1_error_vs_norm} seems an invalid step.
Still, in the limit of large $\alpha$, Eq.~\ref{eq:phi} implies that the whole learning curve has the same power-law scaling with $\alpha$, and therefore we can use Eq.~\ref{eq:scaling_law_2_optnorm_vs_data} for any $\lambda$.
Plugging Eq.~\ref{eq:scaling_law_2_optnorm_vs_data} in Eq.~\ref{eq:scaling_law_1_error_vs_norm} we obtain
\begin{equation}
\hat\epsilon_\mathrm{gen}(\alpha) = k_1\bigl(k_2\alpha^{\gamma_2} + q_2)^{-\gamma_1} + q_1.
\label{eq:hestness}
\end{equation}
For the perceptron Eq.~\ref{eq:hestness} simplifies to $\hat\epsilon_\mathrm{gen}(\alpha) \sim \alpha^{-\gamma_1 \gamma_2}$, and we can recover $\gamma$ as $\gamma_1 \gamma_2$.   For the fixed-norm perceptron we obtain $\gamma_1=-1/2$ (Fig.~\ref{fig:twoScalingLaws}, left panel) and $\gamma_2=1$ (Fig.~\ref{fig:twoScalingLaws}, right panel, upper inset), which recovers $\gamma_1 \gamma_2=\gamma=-1/2$ (Fig.~\ref{fig:twoScalingLaws}, right panel, lower inset). Exponents computed for free-norm perceptron are $\gamma_1=0.4901\pm0.0005$ and $\gamma_2=0.96\pm0.25$; we are unable to estimate $\gamma$ in the free-norm case because training at large $\alpha$ and $\lambda$ requires a number of gradient descent steps that is exponential in $\lambda$ \citep{Soudry2018}. In Appendix \ref{sec:app_perc} we provide analtical arguments to obtain the exponent in the fixed norm case  and describe the numerical methods that we used to compute exponents in both cases.

\begin{figure}
    \centering
    \includegraphics[width=.99\linewidth]{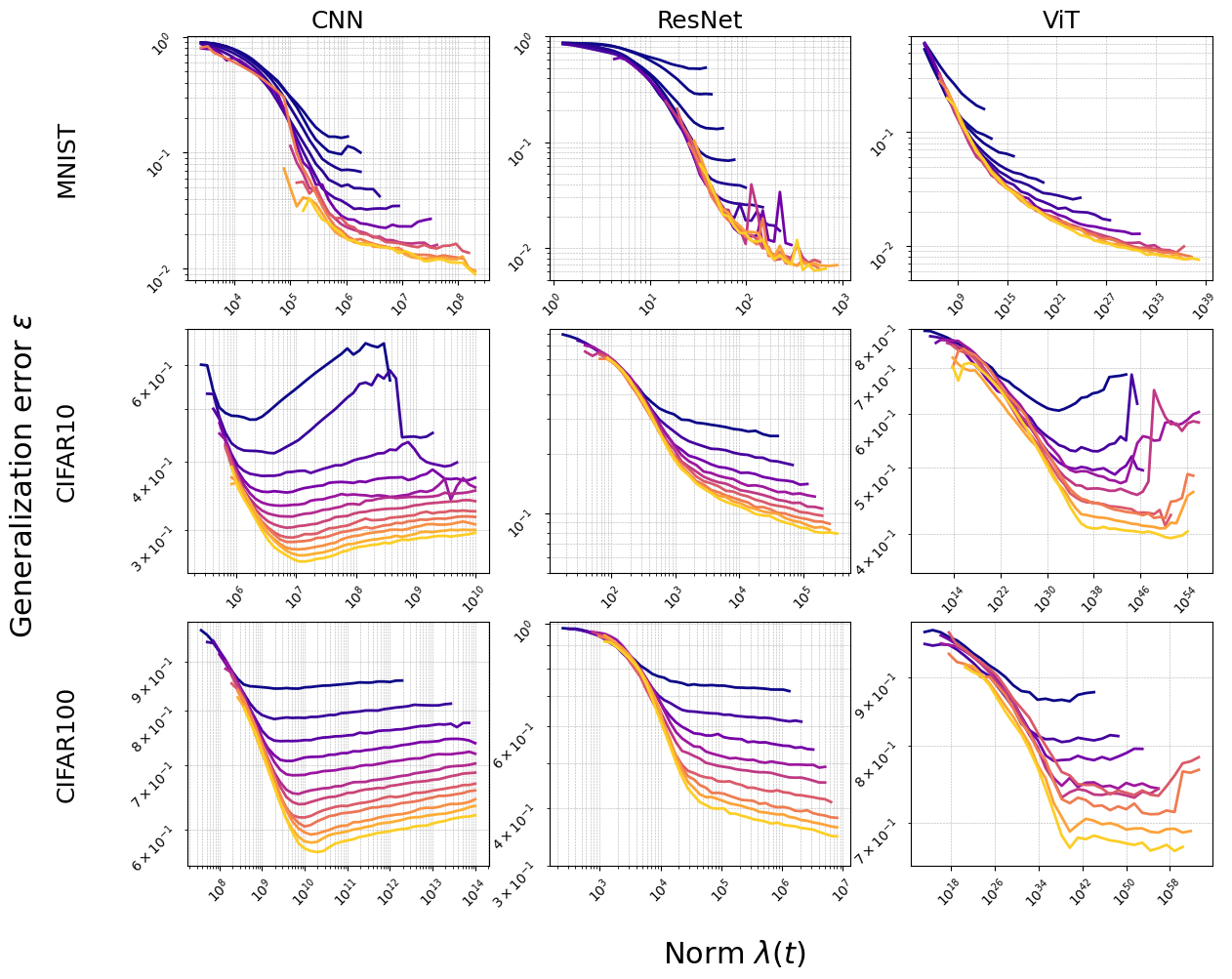}
    \caption{\textbf{Early-training learning curves collapse into a power law when plotted as a function of the spectral complexity norm.} We plot the generalization error $\epsilon$ as a function of the norm $\lambda(t)$ for different datasets and model architectures. Different colors in the same panel refer to training curves with increasing values of the dataset size $P$, ranging from small (blue tones) to large (orange tones). The specific values of $P$ used for each dataset-model combination are listed in Appendix~\ref{sec:spec}.}
    \label{fig:eps_vs_lambda}
\end{figure}

\section{Scaling laws in learning curves of deep architectures}
\label{sec:deep}

\paragraph*{Methods.}
Motivated by results on perceptrons, we repeat for deep architectures the analysis of the test error $\epsilon$ versus increasing norm during training $\lambda(t)$. We test a simple CNN model (that in the following we will simply call "CNN") \citep{lecun1998gradient}, ResNet \citep{he2016residual} and Vision Transformer \citep{dosovitskiy2021image} architectures for image classification over MNIST \citep{lecun1998mnist}, CIFAR10 and CIFAR100 \citep{krizhevsky2009cifar10} datasets.
For each dataset and architecture we make a standard choice of hyperparameters (see Appendix~\ref{sec:spec}), without using a weight decay. Results with moderate weight-decay are reported in Appendix \ref{sec:WD}. For each experiment, we select a random subset of $P$ elements from training set and we train for a fixed number of epochs, large enough to see the test error overfit or saturate. We do this procedure for all values of $P$ selected and then we repeat the training a number of times varying the random subset and of the initial condition of the training. See Appendix \ref{sec:spec} for more details.

\begin{figure}
    \centering
    \includegraphics[width=.99\linewidth]{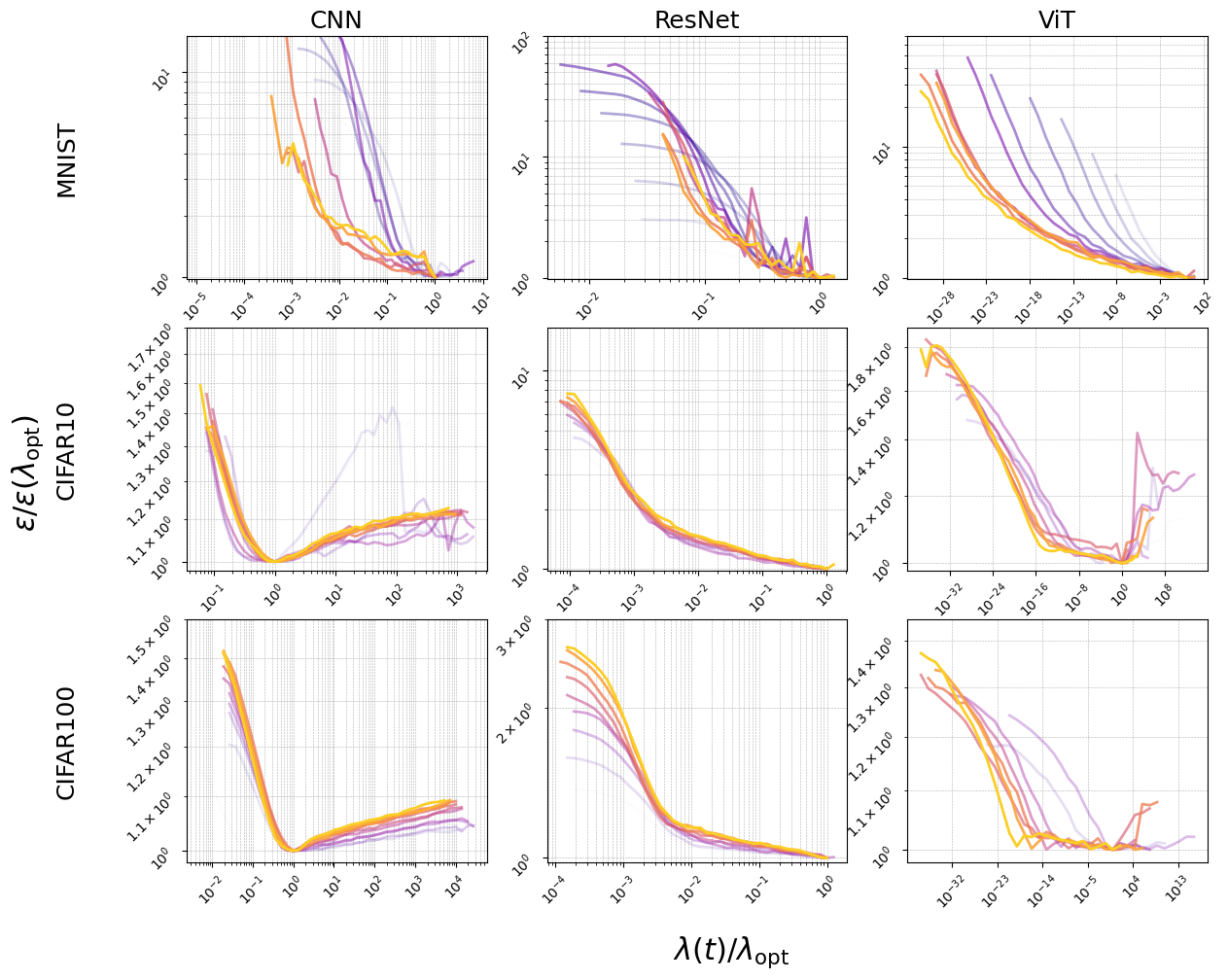}
    \caption{\textbf{The whole learning curves collapse at large $P$ with the proper scalings.} We plot the generalization error $\epsilon$ as a function of the norm $\lambda(t)$ for different datasets and model architectures, rescaling each curve by its optimal point $(\lambda_{\text{opt}}(\alpha), \epsilon_{\text{opt}}(\alpha))$. Different colors in the same panel refer to training curves with increasing values of the dataset size $P$, ranging from small (blue tones) to large (orange tones). The values of $P$ used for each dataset-model combination are listed in Appendix~\ref{sec:spec}.}
    \label{fig:phi}
\end{figure}

For the norm definition in the case of deep networks, in the main analysis we opt for the spectral complexity defined in \cite{Bartlett2017}, 
In that work, the authors show that this quantity has desirable properties for a norm, such as yielding a converging margin distributions that reflect the complexity of the dataset. In  Sec. \ref{sec:results_deep}.
Given the set $A$ of weight matrices $A_i$, the spectral complexity norm $R_A$ of the models reads
\begin{equation}
    R_A = 
\Biggl(\prod_{i=1}^L \rho_i \,\|A_i\|_\sigma\Biggr)
\Biggl(\sum_{i=1}^L 
    \frac{\|A_i^\top - M_i^\top\|_{2,1}^{\,2/3}}
         {\|A_i\|_\sigma^{\,2/3}}
\Biggr)^{3/2},
\label{eq:Bartlett2017}
\end{equation}
where L is the total number of layers in the network, $\rho_i$ is the Lipschitz constant of the activation function (e.g. for ReLU: $\rho_i=1$), $A_i$  is the linear operator at layer $i$ for dense layers and it is an appropriate matrix for convolutional layers (see \cite{Bartlett2017} for a complete explanation). The so-called \textit{reference matrix} $M_i$ is chosen as $0$ for linear or convolutional layers and as the identity for residual layers.
Then, $\|A_i\|_\sigma$ is defined as the largest singular value of $A_i$ and $\|A\|_{2,1}$ is defined as the average of the $\ell_2$-norms of the column vectors. 

Throughout rest of the paper, when we write $\lambda(t)$ for deep architectures we mean the spectral complexity norm $R_{A(t)}$, measured after $t$ training epochs.
We can give an intuition on Eq. \ref{eq:Bartlett2017} by analyzing the contribution of the two terms. Given a layer $i$, first term is the maximum amount that an input vector can be expanded in the output space, and second term is a correction that estimates the effective rank of the outputs of the layer, that is the number of columns that have weights substantially different from zero.
In Appendix~\ref{sec:norm_vs_time} (Fig.~\ref{fig:norm_vs_time}) we show that the relation between $\lambda$ and $t$ is non trivial, and that simply plotting $\epsilon(t)$ does not reveal the same scalings that plotting $\epsilon(\lambda(t))$ does. We always observe the monotonicity of $\lambda(t)$ if a weight-decay is not present. 

\begin{figure}
  \begin{minipage}[b]{0.45\textwidth}
    \centering
    \includegraphics[width=\linewidth]{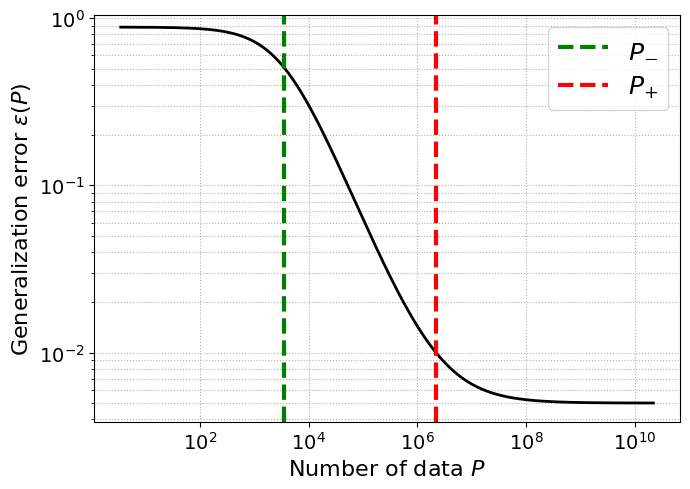}
    \caption{\textbf{The combination of two power laws reproduces known scalings.} We plot the combined power-law scaling of the generalization error as a function of the number of data (Equation \eqref{eq:hestness}). The parameters of the power-law are chosen of the same order of magnitude as typical results obtained for deep networks.}
    \label{fig:empirical}
  \end{minipage}
  \hfill
  \begin{minipage}[b]{0.52\textwidth}
        \begin{tabular}{llccc}
        \toprule
        Model & Dataset & $\gamma_{pred}$ & $\gamma_{meas}$ & $\sigma$     \\
        \midrule
CNN & MNIST & 0.60 & 0.55 & 0.09 \\
CNN & CIFAR10 & 0.28 & 0.25 & 0.07 \\
CNN & CIFAR100 & 0.16 & 0.16 & 0.03 \\
ResNet & MNIST & 0.57 & 0.69 & 0.08 \\
ResNet & CIFAR10 & 0.54 & 0.56 & 0.04 \\
ResNet & CIFAR100 & 0.31 & 0.37 & 0.03 \\
ViT & MNIST & 0.47 & 0.54 & 0.03 \\
ViT & CIFAR10 & 0.23 & 0.21 & 0.03 \\
ViT & CIFAR100 & 0.14 & 0.12 & 0.04 \\
        \bottomrule
        \end{tabular}
    
    \captionof{table}{\textbf{Predicted vs. measured $\epsilon(P)$ exponents across datasets and architectures.}} The exponent $\gamma_{\mathrm{pred}}$ is computed by independently fitting $\gamma_1$ and $\gamma_2$, and combining them as $\gamma_{\mathrm{pred}} = \gamma_1 \gamma_2$. The exponent $\gamma_{\mathrm{meas}}$ is obtained by fitting the $\epsilon(P)$ curves directly. The value of $\sigma$ represents an estimate of the variability of the overall process (see Appendix~\ref{sec:Hest_emp} for details).
    \label{tab:error_exponents}
  \end{minipage}   
\end{figure}

\paragraph{Main result 1: Dynamical scaling laws.}
In this section we consider three architectures without changing their sizes, so $P$ and $\alpha$ are interchangeable. 
In Fig. \ref{fig:eps_vs_lambda} we report the learning curves mediated over different runs, for many values $P$ (the values change for each datasets and are reported in Appendix \ref{sec:spec}, together with the other details of the training process). 
Notably, \textit{we find the same dynamical scaling laws that we observed for perceptrons}: the learning curves have an early training regime independent on $P$ and a late training regime which depends on $P$ (compare  Fig. \ref{fig:eps_vs_lambda} to the left panel of Fig. \ref{fig:twoScalingLaws}). 
In Fig. \ref{fig:phi} we rescale the learning curves in the same way we did for perceptrons (dividing the axes by the optimal norm and optimal error) fiding that they collapse for large $P$ (compare Fig. \ref{fig:phi} to the right panel of  Fig. \ref{fig:twoScalingLaws}). 
Note that, at variance with perceptrons, it is sufficient to plot the generalization error to reveal the scaling laws (and not the relative error).
We stress that the collapse of the learning curves is surprising because we are far from the regime where $P$ is effectively infinite (since increasing $P$ still decreases the generalization error of models): we have a different loss landscape for each value of $P$, and the early stage curves at large $P$ includes the early stage curve of all loss landscapes at lower $P$.

\paragraph*{Main result 2: Recovery of scaling laws at convergence.}
\label{sec:results_deep}
A natural question is wether we can use the measured values of $\gamma_1$ and $\gamma_2$ to recover the end-of-training scaling law in Eq.~\ref{eq:hestness} also for deep models. In this case, since $q_2\neq0$ in general, we need to isolate the proper power law regime. As we show in the sketch in Fig.~\ref{fig:empirical}, it is possible to identify two thresholds
$P_{-} \sim ({q_2}/{k_2})^{1/\gamma_2}$ and
$P_{+} \sim ({k_1 k_2^{-\gamma_1}}/{q_1})^{1/(\gamma_1\gamma_2)}$,
which distinguish between three regimes: 
1) $P\ll P_{-}$, where $\epsilon(P)\,\simeq\,k_1q_2^{-\gamma}+q_1$. In this regime, we expect $\epsilon(P)$ to be close to random guessing, which for classification is $k_1q_2^{-\gamma}+q_1 = (n-1)/n$, with $n$ the number of classes.
2) $P_{-}\ll P\ll P_{+}$, where $\epsilon(P)\,\simeq\,k_1k_2^{-\gamma_1}\,P^{-\gamma_1\gamma_2}$. The exponent $\gamma = \gamma_1\gamma_2$ corresponds to the neural scaling law observed in \cite{Hestness2017}.
3) $P\gg P_{+}$, where $\epsilon(P)\,\to\,q_{1}.$ Here we approach the lowest possible error of the dataset and the performance saturates.

For each architecture and dataset we consider, we measure $\gamma_1,\gamma_2$ with a procedure described in Appendix \ref{sec:Hest_emp} (see the results in Tab.~\ref{tab:error_exponents_2},  Appendix \ref{sec:Hest_emp}). In analogy with perceptrons, we can recover the exponent of end of training scaling law as $\gamma_\mathrm{pred}=\gamma_1\gamma_2$, and compare it to the value $\gamma_\mathrm{meas}$  that we fit directly from the minima of the learning curves at different values of $P$. We observe from Tab. \ref{tab:error_exponents} that in all cases the two values are compatible within the accuracy permitted by the fitting procedure. See Fig. \ref{fig:Hestness_emp} in Appendix \ref{sec:Hest_emp} for a more detailed comparison.

\paragraph{Effect of regularizations, alternative optimizers and different norms.} In Appendix \ref{sec:WD} we show that the qualitative picture of scaling laws in learning curves holds also in the presence of a moderate weight decay. Exponents $\gamma_1$ and $\gamma_2$ change depending on the amount of weight decay, but the values of $\gamma_\mathrm{pred}$ remain compatible within errors with the case without weight decay. In Appendix \ref{sec:SGD} we show that using SGD optimizer instead of Adam in CNN architecture changes the dynamical learning curves, and consequently we obtain different values $\gamma_1$ and $\gamma_2$. However, they produce the same end-of-training exponents $\gamma_\text{pred}=\gamma_1\gamma_2$ as in the main analysis by using Adam. In short, we reproduce the scaling law from \cite{Hestness2017} even when we employ weight decay and an alternative optimizer. In Appendix \ref{sec:other_norms} we show that also four other notions of norm reproduce the qualitative picture of the two scaling laws, but they all find incompatible values of $\gamma_{\text{pred}}$ and $\gamma_{\text{meas}}$, suggesting that only the spectral complexity norm properly captures the scaling behavior.

\section{Discussion}
\label{sec:discussion}

\paragraph{Summary of results.}


Inspired by the implicit bias in perceptrons trained with logistic loss, 
our study uncovers new neural scaling laws in deep architectures that govern how test error evolves throughout training, not just at convergence. 
\begin{itemize}
    \item In perceptrons, we observe that \textbf{the whole learning curve is biased towards specific solutions}.
    Early in the training the perceptron implements Hebbian learning, then it reaches a Bayes-optimal solution and finally it overfits by approaching max-stability rule.
    \item The key point that we learn from perceptrons is to plot the learning curves as function of the increasing norm (we use the spectral-complexity norm for deep architectures).
    The resulting learning curves show two distinct regimes: an \textbf{early-training regimes that follows a power law} that is independent of the size of the training set, and a late-training regime that depends on the size of the training set.
    \item In deep networks, when the \textit{whole} curves are rescaled by the optimal model norm 
    and the corresponding minimum test error, 
    \textbf{learning trajectories from different large-dataset regimes collapse onto a single curve}.
    \item Together, these scaling laws recover the classic end-of-training scaling of test error with data.
\end{itemize}

\paragraph{Possible implications.} The analogies between the scaling laws of perceptrons and deep architectures suggests an implicit bias throughout the whole learning procedure also for deep architectures.  
Overfitting can be seen as follows: although the asymptotic solution maximizes classification margins, the learning trajectory may pass near solutions with fixed spectral complexity and better generalization (cf. perceptrons, Fig.~\ref{fig:boundedVsUnbounded}). An interesting future research line could be to train a deep architecture while constraining its spectral complexity to follow a predetermined trend over time $\lambda(t)$ and study if such training procedure would produce the same learning curves $\epsilon(\lambda)$ .
A second view comes from the self-similarity of early learning: the process first finds a simple solution (low complexity), then gradually increases the norm until reaching the maximum allowed by the dataset size. This provides a pictorial explanation of implicit bias: trajectories with larger datasets shadow those of smaller ones, until late training where overfitting may arise.
A third, practical implication comes from the collapse of learning curves over an asymptotic master curve (Fig. \ref{fig:phi}): it is possible to measure the shape of the generalization error curve on small dataset and predict the same shape for larger datasets, which can be of great practical convenience. However, this method requires a validation with an extensive analysis of robustness across models and datasets, as done for instance in \citet{rosenfeld2019constructive}.

\paragraph{Limitations of the comparison between perceptrons and deep architectures} 
The idea of a training-time bias for perceptrons is fascinating, but in this work it remains mainly qualitative. To obtain quantitative guarantees, one would need an approach similar to \cite{wu_benefits_2025}, or alternatively a full solution of the training dynamics using dynamical mean-field theory (see for example \cite{montanari_dynamical_2025}). Extending these ideas to deep architectures is compelling, but while in perceptrons we can access analytically solutions at fixed norm, there is no obvious analogous picture for deep architectures. The spectral complexity norm seems a good candidate, but the extent to which this property can be made quantitative is unknown.
Here we provide a possible intuition for the success of the comparison between perceptrons and deep models: the classification margin $\Delta$ enters the cross-entropy loss in both in perceptrons (where it is normalized by the L2 norm of weights) and in deep networks.
In \citet{Bartlett2017}, it was shown that spectral complexity norm reproduces the "correct"  normalization of margins in deep architectures. This may be the reason why the spectral complexity norm reveals in deep architectures the same scaling laws of perceptrons
(see \citet{Bartlett2017} for a more detailed definition of "correct").

\paragraph{Limitations and possible extensions of our numerical analysis.}
The main shortcoming of our analysis is that experiments were limited to image classification. We made this choice because we wanted to form a clean conceptual picture before addressing other domains, such as language models, that require larger-scale experiments.
For similar reasons we did not vary the number of parameters for each architecture, limiting our experiments to few standard architectures. 
Moreover, our new scaling laws are motivated by the comparison with a simple and fully understood model, and we lacked a similarly well-understood model for multi-layer perceptrons (in perceptrons we cannot increase arbitrarily the number of parameters because everything depends on the ratio $P/N$ and there is no hidden layer). Recently, some promising works \cite{montanari_dynamical_2025, barbier_statistical_2025}, and we are optimistic that our analysis can be extended in the near future.
Extending our analysis to the joint scaling with width and depth will be essential to understand how our result may impact compute-optimal predictions \citep{Kaplan2020,Henighan2020,Hoffmann2022} (especially in larger models, where these predictions are vital). We expect this direction to be particularly promising, since the spectral complexity norm scales properly with the width and depth of architectures.
Moreover, varying the number of parameters will clarify the role of overparametrization in escaping early-training plateaus, as suggested in \cite{arnaboldi_escaping_2024}.

\paragraph{Final remarks.}
In this work we consolidate the evidence of dynamical scaling laws consistently across dataset and architectures. At the same time, by linking implicit optimization bias with empirical scaling laws, we propose a  picture in which norm growth is the variable that controls neural scaling laws during training. Our findings suggest that the same implicit bias that drives gradient descent toward solutions with maximum margins may also shape the learning trajectory throughout the entire training process, potentially providing a new theoretical framework to understand the emergence of neural scaling laws, and possibly connecting with dynamical scaling laws obtained with other methods \citep{velikanov2021explicit, bordelon2024dynamical, arnaboldi_escaping_2024, montanari_dynamical_2025}.

\paragraph{Acknowledgements}
We thank Brandon Livio Amnesi and Chiara Cammarota for insightful discussions. FD thanks Universidad Complutense de Madrid for its hospitality during his stay, during which part of this work was conducted. This research has been supported by FIS (Italian Science Fund) 2021 funding scheme (FIS783 - SMaC - Statistical Mechanics and Complexity) from MUR, Italian Ministry of University and Research. FD acknowledges funding from the Bando Ricerca Scientifica 2025 - Avvio alla Ricerca (D. R. 2155/2025) of Sapienza Università di Roma, project B83C25004300005 - VESTA. MN aknowledges the support of the PNRR project PE0000013-FAIR, funded by the European Union - NextGenerationEU. This study was conducted using the DARIAH HPC-AI cluster at CNR-NANOTEC in Lecce, funded by the "MUR PON Ricerca e Innovazione 2014-2020" project, code PIR01\_00022 and H2IOSC Project - Humanities and cultural Heritage Italian Open Science Cloud funded by the European Union – NextGenerationEU – NRRP M4C2 - Project code IR0000029. 

\bibliography{refs}

@article{Montanari2024NegativePerceptron,
  author  = {Montanari, Andrea and Zhong, Yiqiao and Zhou, Kangjie},
  title   = {Tractability from Overparametrization: The Example of the Negative Perceptron},
  journal = {Probability Theory and Related Fields},
  year    = {2024},
  volume  = {188},
  number  = {3--4},
  pages   = {805--910},
  doi     = {10.1007/s00440-023-01248-y},
  url     = {https://doi.org/10.1007/s00440-023-01248-y},
  note    = {arXiv:2110.15824}
}

@article{Hestness2017,
  title={Deep learning scaling is predictable, empirically},
  author={Hestness, Joel and Narang, Sharan and Ardalani, Newsha and Diamos, Gregory and Jun, Heewoo and Kianinejad, Hassan and Patwary, Md Mostofa Ali and Yang, Yang and Zhou, Yanqi},
  journal={arXiv preprint arXiv:1712.00409},
  year={2017}
}

@article{Kaplan2020,
  title={Scaling laws for neural language models},
  author={Kaplan, Jared and McCandlish, Sam and Henighan, Tom and Brown, Tom B and Chess, Benjamin and Child, Rewon and others},
  journal={arXiv preprint arXiv:2001.08361},
  year={2020}
}

@article{Henighan2020,
  title={Scaling laws for autoregressive generative modeling},
  author={Henighan, Tom and Kaplan, Jared and Katz, Mor and Chen, Mark and Hesse, Christopher and Jackson, Jacob and Jun, Heewoo and Brown, Tom B and Dhariwal, Prafulla and Gray, Scott and others},
  journal={arXiv preprint arXiv:2010.14701},
  year={2020}
}

@article{velikanov2021explicit,
  title={Explicit loss asymptotics in the gradient descent training of neural networks},
  author={Velikanov, Maksim and Yarotsky, Dmitry},
  journal={Advances in Neural Information Processing Systems},
  volume={34},
  pages={2570--2582},
  year={2021}
}

@article{bordelon2024dynamical,
  title={A dynamical model of neural scaling laws},
  author={Bordelon, Blake and Atanasov, Alexander and Pehlevan, Cengiz},
  journal={arXiv preprint arXiv:2402.01092},
  year={2024}
}

@article{boopathy2024unified,
  title={Unified Neural Network Scaling Laws and Scale-time Equivalence},
  author={Boopathy, Akhilan and Fiete, Ila},
  journal={arXiv preprint arXiv:2409.05782},
  year={2024}
}

@article{Hoffmann2022,
  title={Training compute-optimal large language models},
  author={Hoffmann, Jordan and Borgeaud, Sebastian and Mensch, Arthur and Buchatskaya, Elena and Cai, Trevor and Rutherford, Eliza and Casas, Diego de Las and Hendricks, Lisa Anne and Welbl, Johannes and Clark, Aidan and others},
  journal={arXiv preprint arXiv:2203.15556},
  year={2022}
}

@article{Neyshabur2014,
  title={In search of the real inductive bias: On the role of implicit regularization in deep learning},
  author={Neyshabur, Behnam and Tomioka, Ryota and Srebro, Nathan},
  journal={arXiv preprint arXiv:1412.6614},
  year={2014}
}

@inproceedings{Zhang2017,
  title={Understanding deep learning requires rethinking generalization},
  author={Zhang, Chiyuan and Bengio, Samy and Hardt, Moritz and Recht, Benjamin and Vinyals, Oriol},
  booktitle={International Conference on Learning Representations (ICLR)},
  year={2017}
}

@inproceedings{sun2017revisiting,
  title={Revisiting unreasonable effectiveness of data in deep learning era},
  author={Sun, Chen and Shrivastava, Abhinav and Singh, Saurabh and Gupta, Abhinav},
  booktitle={Proceedings of the IEEE international conference on computer vision},
  pages={843--852},
  year={2017}
}

@article{rosenfeld2019constructive,
  title={A constructive prediction of the generalization error across scales},
  author={Rosenfeld, Jonathan S and Rosenfeld, Amir and Belinkov, Yonatan and Shavit, Nir},
  journal={arXiv preprint arXiv:1909.12673},
  year={2019}
}

@article{Soudry2018,
  title={The implicit bias of gradient descent on separable data},
  author={Soudry, Daniel and Hoffer, Elad and Nacson, Mor Shpigel and Gunasekar, Suriya and Srebro, Nathan},
  journal={Journal of Machine Learning Research},
  volume={19},
  number={70},
  pages={1--57},
  year={2018}
}

@inproceedings{Gunasekar2017,
  title={Implicit regularization in matrix factorization},
  author={Gunasekar, Suriya and Woodworth, Blake and Bhojanapalli, Srinadh and Neyshabur, Behnam and Srebro, Nathan},
  booktitle={Advances in Neural Information Processing Systems},
  volume={30},
  pages={6151--6159},
  year={2017}
}

@inproceedings{Lyu2020,
  title={Gradient descent maximizes the margin of homogeneous neural networks},
  author={Lyu, Kaifeng and Li, Jian},
  booktitle={International Conference on Learning Representations (ICLR)},
  year={2020}
}

@inproceedings{Chizat2020,
  title={Implicit bias of gradient descent for wide two-layer neural networks trained with the logistic loss},
  author={Chizat, Lénaïc and Bach, Francis},
  booktitle={Conference on Learning Theory (COLT)},
  year={2020}
}

@book{Engel2001,
  title={Statistical Mechanics of Learning},
  author={Engel, Andreas and Van den Broeck, Christian},
  year={2001},
  publisher={Cambridge University Press}
}

@inproceedings{Bartlett2017,
  title={Spectrally-normalized margin bounds for neural networks},
  author={Bartlett, Peter L and Foster, Dylan J and Telgarsky, Matus J},
  booktitle={Advances in Neural Information Processing Systems},
  volume={30},
  pages={6240--6249},
  year={2017}
}

@book{mezard1987spin,
  title={Spin glass theory and beyond: An Introduction to the Replica Method and Its Applications},
  author={M{\'e}zard, Marc and Parisi, Giorgio and Virasoro, Miguel Angel},
  volume={9},
  year={1987},
  publisher={World Scientific Publishing Company}
}

@article{lecun1998gradient,
  title        = {Gradient-based Learning Applied to Document Recognition},
  author       = {LeCun, Yann and Bottou, L\'eon and Bengio, Yoshua and Haffner, Patrick},
  journal      = {Proceedings of the IEEE},
  volume       = {86},
  number       = {11},
  pages        = {2278--2324},
  year         = {1998},
  doi          = {10.1109/5.726791},
  url          = {https://vision.stanford.edu/cs598_spring07/papers/Lecun98.pdf}
}

@inproceedings{he2016residual,
  title        = {Deep Residual Learning for Image Recognition},
  author       = {He, Kaiming and Zhang, Xiangyu and Ren, Shaoqing and Sun, Jian},
  booktitle    = {Proceedings of the IEEE Conference on Computer Vision and Pattern Recognition (CVPR)},
  pages        = {770--778},
  year         = {2016},
  month        = jun,
  publisher    = {IEEE},
  doi          = {10.1109/CVPR.2016.90},
  url          = {http://ieeexplore.ieee.org/document/7780459}
}

@inproceedings{dosovitskiy2021image,
  title        = {An Image is Worth 16x16 Words: Transformers for Image Recognition at Scale},
  author       = {Dosovitskiy, Alexey and Beyer, Lucas and Kolesnikov, Alexander and Weissenborn, Dirk and Zhai, Xiaohua and Unterthiner, Thomas and Dehghani, Mostafa and Minderer, Matthias and Heigold, Georg and Gelly, Sylvain and Uszkoreit, Jakob and Houlsby, Neil},
  booktitle    = {International Conference on Learning Representations (ICLR)},
  year         = {2021},
  url          = {https://arxiv.org/abs/2010.11929}
}

@misc{lecun1998mnist,
  title        = {{MNIST} Handwritten Digit Database},
  author       = {LeCun, Yann and Cortes, Corinna and Burges, Christopher J. C.},
  howpublished = {\url{http://yann.lecun.com/exdb/mnist/}},
  year         = {1998},
  note         = {Accessed: 2025-05-14}
}

@techreport{krizhevsky2009cifar10,
  title        = {Learning Multiple Layers of Features from Tiny Images (CIFAR-10 Dataset)},
  author       = {Krizhevsky, Alex and Hinton, Geoffrey},
  institution  = {University of Toronto},
  year         = {2009},
  number       = {Technical Report 0},
  url          = {https://www.cs.toronto.edu/~kriz/learning-features-2009-TR.pdf}
}

@misc{wu_benefits_2025,
	title = {Benefits of {Early} {Stopping} in {Gradient} {Descent} for {Overparameterized} {Logistic} {Regression}},
	url = {http://arxiv.org/abs/2502.13283},
	doi = {10.48550/arXiv.2502.13283},
	language = {en},
	urldate = {2025-07-10},
	publisher = {arXiv},
	author = {Wu, Jingfeng and Bartlett, Peter and Telgarsky, Matus and Yu, Bin},
	month = jun,
	year = {2025},
	note = {arXiv:2502.13283 [cs]},
}

@misc{arnaboldi_escaping_2024,
	title = {Escaping mediocrity: how two-layer networks learn hard generalized linear models with {SGD}},
	shorttitle = {Escaping mediocrity},
	url = {http://arxiv.org/abs/2305.18502},
	doi = {10.48550/arXiv.2305.18502},
	urldate = {2025-09-21},
	publisher = {arXiv},
	author = {Arnaboldi, Luca and Krzakala, Florent and Loureiro, Bruno and Stephan, Ludovic},
	month = mar,
	year = {2024},
	note = {arXiv:2305.18502 [stat]},
}

@misc{montanari_dynamical_2025,
	title = {Dynamical {Decoupling} of {Generalization} and {Overfitting} in {Large} {Two}-{Layer} {Networks}},
	url = {http://arxiv.org/abs/2502.21269},
	doi = {10.48550/arXiv.2502.21269},
	urldate = {2025-09-21},
	publisher = {arXiv},
	author = {Montanari, Andrea and Urbani, Pierfrancesco},
	month = sep,
	year = {2025},
	note = {arXiv:2502.21269 [stat]},
}

@inproceedings{aubin_generalization_2020,
	title = {Generalization error in high-dimensional perceptrons: {Approaching} {Bayes} error with convex optimization},
	volume = {33},
	shorttitle = {Generalization error in high-dimensional perceptrons},
	url = {https://proceedings.neurips.cc/paper/2020/hash/8f4576ad85410442a74ee3a7683757b3-Abstract.html},
	urldate = {2025-06-09},
	booktitle = {Advances in {Neural} {Information} {Processing} {Systems}},
	publisher = {Curran Associates, Inc.},
	author = {Aubin, Benjamin and Krzakala, Florent and Lu, Yue and Zdeborová, Lenka},
	year = {2020},
	pages = {12199--12210},
}

@article{gardner1987capacity,
  title={Maximum storage capacity in neural networks},
  author={Gardner, Elizabeth},
  journal={Europhysics Letters},
  volume={4},
  number={4},
  pages={481--485},
  year={1987},
  publisher={IOP Publishing},
  doi={10.1209/0295-5075/4/4/004}
}

@article{gardner1988optimal,
  title={Optimal storage properties of neural network models},
  author={Gardner, Elizabeth and Derrida, Bernard},
  journal={Journal of Physics A: Mathematical and general},
  volume={21},
  number={1},
  pages={271},
  year={1988},
  publisher={IOP Publishing}
}

@article{opper1988learning,
  title={Learning times of neural networks: Exact solution for a perceptron algorithm},
  author={Opper, Manfred},
  journal={Physical Review A},
  volume={38},
  number={8},
  pages={3824--3826},
  year={1988},
  publisher={APS},
  doi={10.1103/PhysRevA.38.3824}
}

@article{opper1990generalization,
  title={On the ability of the optimal perceptron to generalize},
  author={Opper, Manfred and Kinzel, Wolfgang and Kleinz, Jan and Nehl, Ralf},
  journal={Journal of Physics A: Mathematical and General},
  volume={23},
  number={11},
  pages={L581--L586},
  year={1990},
  publisher={IOP Publishing},
  doi={10.1088/0305-4470/23/11/003}
}

@article{opper1991bayes,
  title={Generalization performance of the optimal Bayes algorithm for learning a perceptron},
  author={Opper, Manfred and Haussler, David},
  journal={Physical Review Letters},
  volume={66},
  number={21},
  pages={2677--2680},
  year={1991},
  publisher={APS},
  doi={10.1103/PhysRevLett.66.2677}
}

@article{biehl1994online,
  title={On-line learning with a perceptron},
  author={Biehl, Michael and Riegler, Peter},
  journal={Europhysics Letters},
  volume={28},
  number={7},
  pages={525--530},
  year={1994},
  publisher={IOP Publishing},
  doi={10.1209/0295-5075/28/7/004}
}

@article{saad1995exact,
  title={Exact solution for on-line learning in multilayer neural networks},
  author={Saad, David and Solla, Sara A.},
  journal={Physical Review Letters},
  volume={74},
  number={21},
  pages={4337--4340},
  year={1995},
  publisher={APS},
  doi={10.1103/PhysRevLett.74.4337}
}

@article{saad1995online,
  title={On-line learning in soft committee machines},
  author={Saad, David and Solla, Sara A.},
  journal={Physical Review E},
  volume={52},
  number={4},
  pages={4225--4243},
  year={1995},
  publisher={APS},
  doi={10.1103/PhysRevE.52.4225}
}

@incollection{solla1998optimal,
  title={Optimal perceptron learning: an on-line Bayesian approach},
  author={Solla, Sara A. and Winther, Ole},
  booktitle={On-line Learning in Neural Networks},
  editor={Saad, David},
  pages={157--178},
  publisher={Cambridge University Press},
  year={1998},
  doi={10.1017/CBO9780511569920.009}
}

@misc{broken_neural,
      title={Broken Neural Scaling Laws}, 
      author={Ethan Caballero and Kshitij Gupta and Irina Rish and David Krueger},
      year={2023},
      eprint={2210.14891},
      archivePrefix={arXiv},
      primaryClass={cs.LG},
      url={https://arxiv.org/abs/2210.14891}, 
}

@misc{emergent_abilities,
      title={Are Emergent Abilities of Large Language Models a Mirage?}, 
      author={Rylan Schaeffer and Brando Miranda and Sanmi Koyejo},
      year={2023},
      eprint={2304.15004},
      archivePrefix={arXiv},
      primaryClass={cs.AI},
      url={https://arxiv.org/abs/2304.15004}, 
}

@misc{survey_scaling_laws,
      title={(Mis)Fitting: A Survey of Scaling Laws}, 
      author={Margaret Li and Sneha Kudugunta and Luke Zettlemoyer},
      year={2025},
      eprint={2502.18969},
      archivePrefix={arXiv},
      primaryClass={cs.LG},
      url={https://arxiv.org/abs/2502.18969}, 
}

@misc{emergent,
      title={Emergent Abilities of Large Language Models}, 
      author={Jason Wei and Yi Tay and Rishi Bommasani and Colin Raffel and Barret Zoph and Sebastian Borgeaud and Dani Yogatama and Maarten Bosma and Denny Zhou and Donald Metzler and Ed H. Chi and Tatsunori Hashimoto and Oriol Vinyals and Percy Liang and Jeff Dean and William Fedus},
      year={2022},
      eprint={2206.07682},
      archivePrefix={arXiv},
      primaryClass={cs.CL},
      url={https://arxiv.org/abs/2206.07682}, 
}

@misc{barbier_statistical_2025,
	title = {Statistical physics of deep learning: {Optimal} learning of a multi-layer perceptron near interpolation},
	url = {http://arxiv.org/abs/2510.24616},
	doi = {10.48550/arXiv.2510.24616},
	publisher = {arXiv},
	author = {Barbier, Jean and Camilli, Francesco and Nguyen, Minh-Toan and Pastore, Mauro and Skerk, Rudy},
	month = oct,
	year = {2025},
    eprint={2510.24616},
    archivePrefix={arXiv},
}
\bibliographystyle{apalike}

\newpage
\appendix

\section*{Appendix}

\paragraph{Acknowledgment of LLMs usage.} The authors acknowledge the usage of LLMs for polishing the text and to produce standard functions in the code for deep networks experiments. All texts and codes produced by LLMs have been carefully analyzed and validated by the authors.

\paragraph{Code to reproduce the results of the paper} All the codes, data, hyperparameters and results on deep architectures can be found in the GitHub repository \url{https://github.com/Francill99/deep_norm.git}.

\section{Replica Analysis}
\label{sec:replicas}
In this section, we provide a sketch of the necessary computations to obtain the analytical curve for the fixed-norm perceptron. We are interested in computing the generalization error, defined as the expected fraction of misclassified examples on new data. In the teacher-student setup for the perceptron presented in the main text, this is given by $\epsilon = \frac{1}{\pi} \text{arccos}(R)$, where $R \equiv (\pmb{w} \cdot \pmb{w}^*)/N$ is the normalized overlap between the student and the teacher.

Given a loss function of the form
\begin{equation}
    L(\pmb{w}) =  \sum_{\mu=1}^{P \equiv \alpha N}V(\Delta^\mu),
    \label{eq:generic_loss}
\end{equation}
where
$
    \Delta^\mu \equiv y^{\mu} \left(\frac{\pmb{w} \cdot \pmb{x}^\mu}{\sqrt{N}}\right)
$ is the \emph{margin} of the $\mu$-th example, we therefore need to compute the typical overlap $\Bar{R}$ between a minimizer of Equation \eqref{eq:generic_loss} and the teacher. To do this, one can study the averaged free energy, defined as
\begin{equation}
\label{eq:free_energy}
f(\beta) = \lim_{N \to \infty} \left( - \frac{1}{\beta N} \left\langle\langle \ln Z \right\rangle\rangle_{\pmb{x}^\mu,\, \pmb{w}^*} \right),
\end{equation}
where $\beta$ is the inverse temperature, $\left\langle \cdot \right\rangle_{\pmb{x}^\mu,\, \pmb{w}^*}$ denotes the average over the distribution of the data points $\{\pmb{x}^\mu\}$ and the teacher vector $\pmb{w}^*$. $Z$ is the partition function defined, as
\begin{equation}
    Z(\pmb{w}) \equiv \int d\mu(\pmb{w})  \; e^{-\beta L(\pmb{w})},
\end{equation}
where $\mu(\pmb{w})$ is the probability distribution of the student vectors, assumed to be uniform on the $N$-sphere. In the thermodynamic limit \( N \to \infty \), only a subset of students, characterized by an overlap with the teacher \(\bar{R}(\beta)\), contributes to \( f(\beta) \). By taking the limit \(\beta \to \infty\), one can obtain the typical overlap considering only the minimizers of the loss.

To compute the average of \( \ln Z \) in Equation \eqref{eq:free_energy}, we apply the replica method \citep{mezard1987spin}, which involves rewriting the logarithmic average as 
\[
\langle \langle \ln Z \rangle \rangle = \lim_{n \to 0} \frac{\langle\langle Z^n \rangle \rangle - 1}{n},
\]
where $Z^n$ is the replicated partition function defined by
\begin{equation}
Z^{(n)} \equiv \left\langle\!\left\langle Z^n(\pmb{x}^\mu, \pmb{w}^*) \right\rangle\!\right\rangle_{\pmb{x}^\mu,\, \pmb{w}^*}
= \left\langle\!\left\langle \int \prod_{a=1}^n d\mu(\pmb{w}^a) \prod_{a=1}^n \exp\left(-\beta L(\pmb{w}^a) \right) \right\rangle\!\right\rangle_{\pmb{x}^\mu,\, \pmb{w}^*}.
\end{equation}

One can introduce new variables 
$R^a = (\pmb{w}^* \cdot \pmb{w}^a)/N$ and $q_{ab} = (\pmb{w}^a \cdot \pmb{w}^b)/N$, which represent the normalized overlap of student $a$ with the teacher, and the overlap between student vectors $a$ and $b$, respectively. The free energy function can then be rewritten in terms of these new variables. Under the replica symmetric ansatz, i.e., choosing solutions of the form
\begin{equation}
    R^a = R \quad \forall a \in [1, n], \quad q_{ab} = \delta_{ab} + q (1 - \delta_{ab}) \quad \forall a,b \in [1, n].
    \label{eq:RS_ansatz}
\end{equation}
one obtains 
\begin{align}
f(\beta) = - \underset{q, R}{\text{extr}} &\left[ \frac{1}{2\beta} \ln(1 - q) + \frac{q - R^2}{2\beta(1 - q)} \right. \notag \\
&\times \left. \ln \int d\Delta \frac{1}{\sqrt{2\pi(1 - q)}} \exp \left( -\beta V(\Delta) - \frac{(\Delta - \sqrt{q}t)^2}{2(1 - q)} \right) \right],
\label{eq:free_energy_RS}
\end{align}
where $H(x) = \frac{1}{2} \, \text{erfc} \left( \frac{x}{\sqrt{2}} \right) = \frac{1}{2} \left( 1 - \text{erf} \left( \frac{x}{\sqrt{2}} \right) \right)$.

If the potential $V(\Delta)$ has a unique minimum, one can evaluate the zero-temperature limit of Equation \eqref{eq:free_energy_RS}, yielding
\begin{align}
f(T = 0) 
= - &\underset{x, R}{\text{extr}} \left[ \frac{1 - R^2}{2x}  - 2\alpha \int \frac{dt}{\sqrt{2\pi}} e^{-t^2/2} \, H\left( -\frac{Rt}{\sqrt{1 - R^2}} \right)\right. \notag\\
&\left. 
\times\left( V(\Delta_0(t,x)) + \frac{(\Delta_0(t,x) - t)^2}{2x} \right) \right] \equiv e(x, R),
\label{eq:free_energy_zeroTemperature}
\end{align}
where $x \equiv \beta(1 - q)$ and 
$\Delta_0(t,x) \equiv \text{argmin}_{\Delta} \left( V(\Delta) + \frac{(\Delta - t)^2}{2x} \right)$. By solving the saddle-point equations
\begin{equation*}
\left. \frac{\partial e}{\partial x} \right|_{x = \bar{x},\, R = \bar{R}} = 0, \quad
\left. \frac{\partial e}{\partial R} \right|_{x = \bar{x},\, R = \bar{R}} = 0,
\end{equation*}
one can finally recover the value $\Bar{R}$ and, consequently, the generalization error.

\section{Perceptron in the over parametrized regime}
\label{sec:app_overparam}
\begin{figure}[H]
    \centering
    \includegraphics[width=0.5\linewidth]{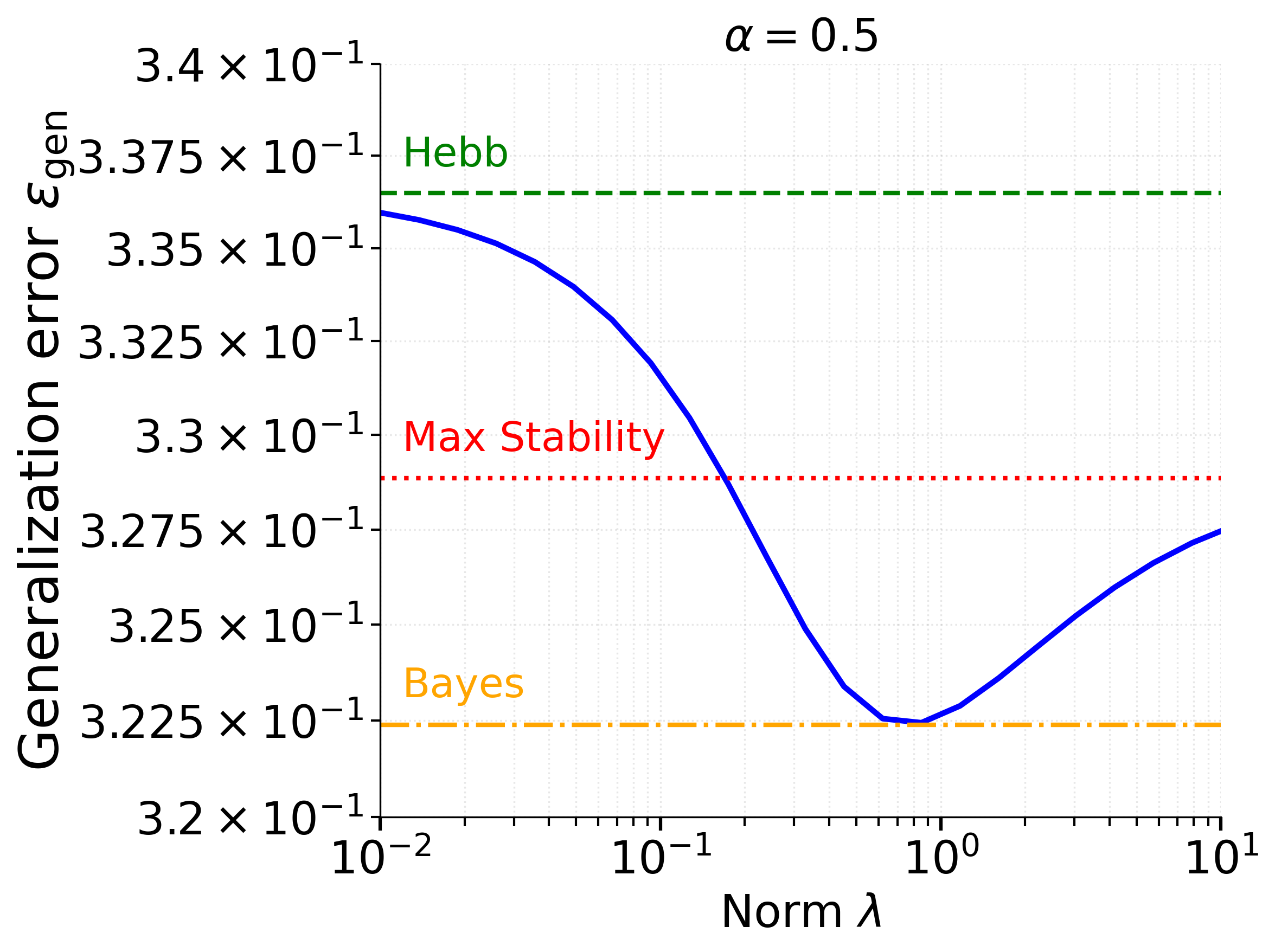}
    \caption{\textbf{The fixed-norm problem is qualitatively the same in the overparametrized regime.} We show the generalization error of the minimizers of the cross-entropy loss in the teacher-student setup for $\alpha = 0.5$,}
    \label{fig:app_overapam}
\end{figure}

In this section we show that the analysis of the different regimes in $\lambda$, shown in the main text for $\alpha>1$, is qualitatively equivalent in the regime $\alpha < 1$. In Figure \ref{fig:app_overapam}, we plot the generalization error as a function of the parameter $\lambda$ for $\alpha = 0.5$.

\section{MSE loss in Perceptron}
\begin{figure}[H]
    \centering
    \includegraphics[width=0.5\linewidth]{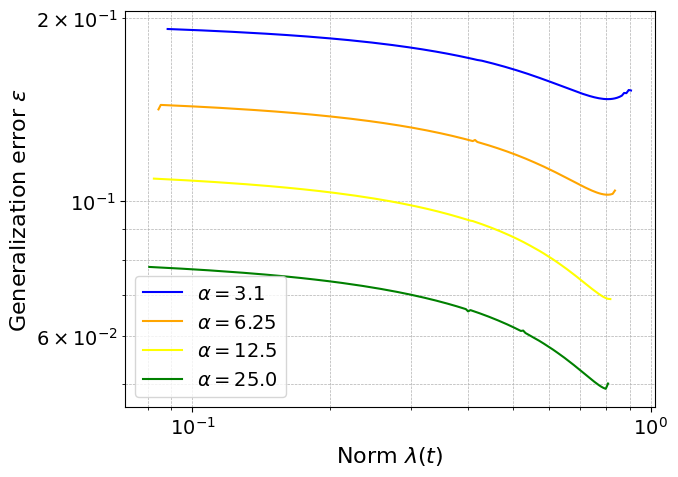}
    \caption{\textbf{MSE loss does not produce the scaling laws in learning curves as Cross-Entropy}. Norm $\lambda(t)$ increases during training epochs up to a certain value, where training stops. We do not observe a scaling law in early training of the form $\epsilon \sim \lambda^{-\gamma_1}$, and $\lambda_\text{opt}(\alpha)$ is constant, not following  second scaling-law.}
    \label{fig:MSE}
\end{figure}

In Fig. \ref{fig:MSE} we show numerical results obtained with the same perceptron settings as in the main analysis with the only difference that loss is chosen as MSE. The qualitative picture is fundamentally different and we do not observe the same phenomenology.
\label{sec:app_MSE}

\section{Analysis of the scaling laws in Perceptron}
\label{sec:app_perc}
\paragraph*{Fixed-norm analytical perceptron} We provide an analytical argument to obtain the first scaling law of the perceptron. We begin from the zero-temperature free energy per neuron:
\[
  e_{\min}(\alpha)
  = -\underset{x>0,\,-1<R<1}{\mathrm{extr}}\Bigl[\frac{1-R^2}{2x}
    -2\alpha\int_{-\infty}^{\infty}Dt\;H\bigl(-R\,t/\sqrt{1-R^2}\bigr)
    \min_{\Delta}\{V_\lambda(\Delta)+\tfrac{(\Delta-t)^2}{2x}\}\Bigr],
\]
where
\(
  Dt=\tfrac{dt}{\sqrt{2\pi}}e^{-t^2/2},
  \quad
  H(u)=\int_u^{\infty}Dt.
\)

Reusing the definition \[
  \Delta_{0,\lambda}(t,x)=\arg\min_{\Delta}\Bigl\{V_\lambda(\Delta)+\tfrac{(\Delta-t)^2}{2x}\Bigr\},
\]
stationarity w.r.t.
\(x\) and
\(R\) yields the coupled equations:
\begin{subequations}
\begin{align}
1 - R^2 &= 2\alpha\int Dt\,(\Delta_{0,\lambda}-t)^2\;H\bigl(-R\,t/\sqrt{1-R^2}\bigr),\label{eq:spx}\\
R &= \frac{2\alpha}{\sqrt{2\pi\,(1-R^2)}}
      \int Dt\;\Delta_{0,\lambda}(t,x)\;\exp\!\Bigl(-\tfrac{R^2t^2}{2(1-R^2)}\Bigr).
      \label{eq:spr}
\end{align}
\end{subequations}
We focus on the regime \(\alpha\to\infty\), where
\(R=1-\delta,\;\delta\ll1\;,x\ll 1\), and the generalization error
\(\varepsilon\equiv1/\pi  \; arccos(R)\approx\sqrt{2\delta}/\pi\). For $x\ll1$, we can solve the equation for $\Delta_0$,
\begin{equation}
    V'(\Delta_0) + \frac{\Delta_0 - t}{x} = 0
\end{equation}
order by order. By assuming that the derivative of the potential is negligible with respect to $1/x$, at first order $\Delta_0 = t$, since $x\ll1$.  We then assume $\Delta_0 \sim t + c\,x$. The minimizing equation leads to $V'(t + c\;x) + c \sim V'(t) + c\;xV''(t) + c=0 \implies c=-V'(t)$, where we have implicitly assumed that $V''(t)$ is negligible respect to $1/x$. At the end we have 
\begin{equation}
\boxed{
    \Delta_{0, \lambda} \sim t -V_\lambda'(t)\;x
    }.
\end{equation}
As \(R\to1\),
\[
  H\bigl(-R\,t/\sqrt{1-R^2}\bigr)\longrightarrow\Theta(t),
  \quad
  \exp\!\Bigl(-\tfrac{R^2t^2}{2(1-R^2)}\Bigr)\longrightarrow\exp\!\Bigl(-\tfrac{t^2}{4\delta}\Bigr).
\]
By plugging these two expressions into \eqref{eq:spx}, we get
\[
  1-R^2 \approx 2\delta,
  \quad
  2\alpha\int_{t>0}Dt\,(\Delta_{0,\lambda}-t)^2\approx2\alpha\,\bigl\langle(\Delta_{0,\lambda}-t)^2\bigr\rangle_{t>0}.
\]
Since  \((\Delta_{0,\lambda}-t)^2\sim x^2 V'_\lambda(t)^2\), we have
\[
  \delta \sim \alpha\,x^2\,\Sigma_0(\lambda),
\]
where we have introduced
\[
  \Sigma_0(\lambda) = \int_{t>0}Dt\,\frac{(\Delta_{0,\lambda}(t,x)-t)^2}{x^2}\,\xrightarrow[x\to0]{}\int_{t>0} Dt\,[V'_\lambda(t)]^2.
\]
Similarly, from \eqref{eq:spr} with the combined Gaussian:
\[
  R \approx \frac{2\alpha}{\sqrt{4\pi\delta}}
    \int Dt\, e^{-t^2/(4\delta)} \Delta_{0,\lambda}(t,x)
  \sim \alpha \int \frac{dt}{\sqrt{2\pi\delta}}\, e^{-t^2/(4\delta)} \left(e^{-t^2/2}\, \Delta_{0,\lambda}(t,x)\right).
\]
This integral exhibits a \emph{delta sequence structure} in the $\delta \to 0$ limit, since the prefactor
\[
\frac{1}{\sqrt{2\pi\delta}}\, e^{-t^2/(4\delta)}
\]
acts as an approximation to the \emph{Dirac delta function} $\delta_D(t)$. Therefore, the integral localizes around $t = 0$, and we obtain:
\[
R \sim \alpha \cdot \Delta_{0,\lambda}(0,x) \sim \alpha \,xV_\lambda'(0)
\]

\noindent Since \(R\approx1\):
\begin{equation}
    \boxed{
  x \sim \frac{1}{\alpha V_\lambda'(0)}.
  }
\end{equation}
Substituting \(x\) into \(\delta\sim\alpha\,x^2\,\Sigma_0(\lambda)\):
\begin{equation}
\boxed{
  \delta \sim \alpha \; \alpha^{-2} (V_\lambda(0))^{-2} \; \Sigma_0(\lambda) = \frac{\Sigma_0(\lambda)}{\alpha V'_\lambda(0)^2}
  }
\end{equation}

For
$\
V_{\lambda}(\Delta) = \Delta - \tfrac{1}{\lambda} \ln[2\cosh(\lambda \Delta)],
$
we compute
\[
V_\lambda'(\Delta) = 1 - \tanh(\lambda \Delta), \quad
V_\lambda'(0) = 1.
\]
We now turn to 
\[
  \Sigma_0(\lambda)=\int_{t>0} Dt\,[V'(t)]^2=\int_{t>0} Dt\,[1-\tanh(\lambda t)]^2.
\]
\noindent For large $\lambda$ we have: 
\[
\tanh(\lambda t) \sim 1 - 2 e^{-2\lambda t} + \ldots.
\]

\noindent Then, at leading order:
\begin{align}
  \Sigma_0(\lambda) &\underset{\lambda \gg 1}{\sim} \int_{t > 0} Dt\, 4 e^{-4\lambda t}
  = 4 \int_{t > 0} \frac{dt}{\sqrt{2\pi}} e^{-t^2/2} e^{-4\lambda t}
  \underset{t' = \lambda t}{=} \frac{4}{\lambda} \int_{t' > 0} \frac{dt'}{\sqrt{2\pi}} e^{-t'^2/(2\lambda^2)} e^{-4 t'} \notag \\
  &\sim \frac{4}{\lambda} \int_{t' > 0} \frac{dt'}{\sqrt{2\pi}} e^{-4t'}
  \sim \frac{C}{\lambda}
\end{align}

\noindent Putting this back into $\delta$, one finally gets for the log cosh potential

\begin{equation}
    \boxed{
     \delta \sim \frac{1}{\alpha \lambda} \implies \varepsilon\sim (\alpha \lambda)^{-1/2}
     }
\end{equation}

In order to understand the regime of validity of this scaling law, we analyze the second derivative of the potential
\[
V_\lambda''(\Delta) = -\lambda\, \text{sech}^2(\lambda \Delta),
\]
so in particular,
\[
V_\lambda''(0) = -\lambda.
\]
This means that the hypothesis $V''(t)x \sim \lambda/\alpha \ll 1$  is no longer valid when $\lambda \sim \alpha$, implying that the regime of validity of this power law is
\begin{equation}
    \boxed{
    1 \ll \lambda \ll \alpha
    }
\end{equation}

\begin{figure}[H]
    \centering
    \includegraphics[width=0.5\linewidth]{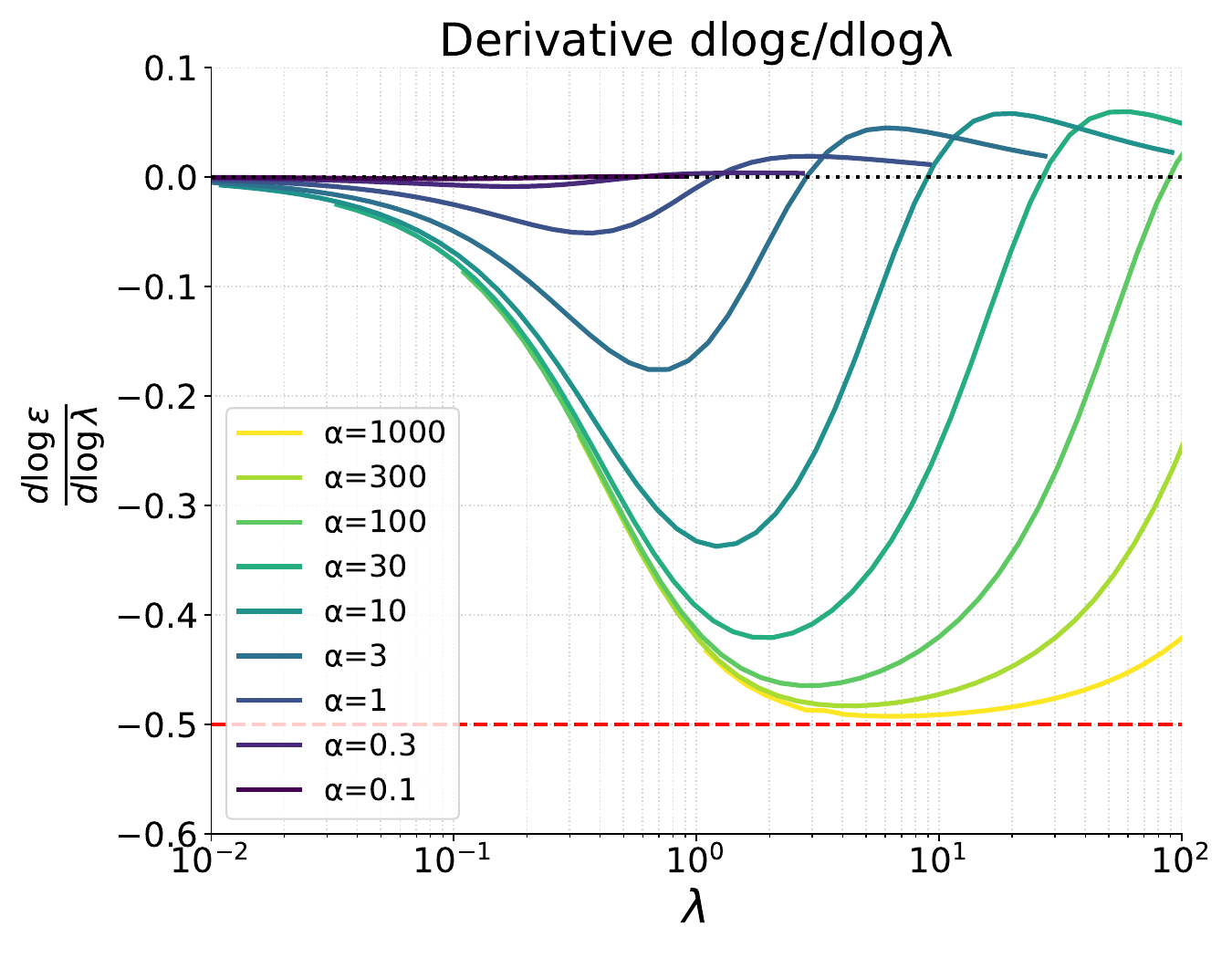}
    \caption{\textbf{Convergence of the perceptron learning exponent.} 
    We plot $\tfrac{d \log \epsilon}{d \log \lambda}$ for different values of $\alpha$. 
    As $\alpha$ increases, an extended region of $\lambda$ develops where the effective exponent approaches $-1/2$, 
    which corresponds to the asymptotic behavior $\epsilon \sim \lambda^{-1/2}$. 
    The dashed red line marks the reference slope $-1/2$ while the black dotted line marks the zero derivative point.}
    \label{fig:exponent_convergence}
\end{figure}

We now provide numerical evidence of the convergence to a $-1/2$ exponent in the $\epsilon(\lambda)$ curve for the perceptron by analyzing the theoretical curves. In Fig.~\ref{fig:exponent_convergence} we plot $\tfrac{d \log \epsilon}{d \log \lambda}$ for different values of $\alpha$, showing that as $\alpha$ increases there appears a broader region of $\lambda$ where the effective exponent approaches $-1/2$.

\paragraph*{Unbounded numerical perceptron}
We compute the two exponents of the unbounded perceptron. For consistency, we have chosen to follow the same procedure that resulted to be the best for deep networks experiments, reported in Appendix \ref{sec:Hest_emp}. In Fig. \ref{fig:fit_perc} we report the fitting plot for $\gamma_1$ and $\gamma_2$ exponents. The two exponents result not compatible considering errors with the analytical result for fixed-norm perceptrons $\gamma_1=0.5, \gamma_2=1.0$, but the differences are only of the order of 5\%. So not only the fixed-norm analytical case predict qualitatively the dynamical behavior of the unbounded perceptron, it also approximates quantitatively the values of the dynamical exponents.
\begin{figure}[H]
    \centering
    \includegraphics[width=0.47\linewidth]{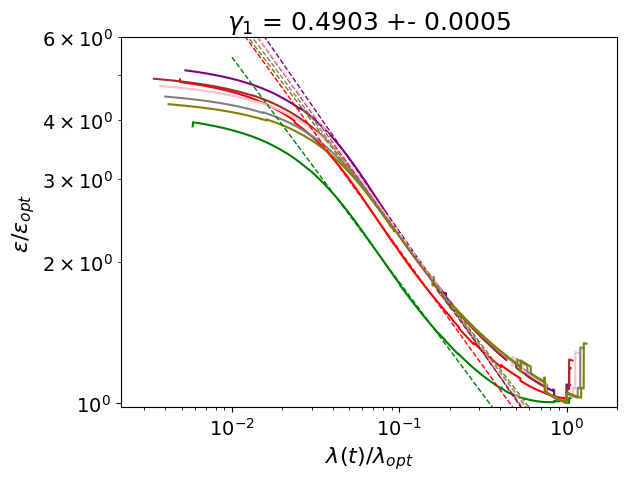}
    \includegraphics[width=0.47\linewidth]{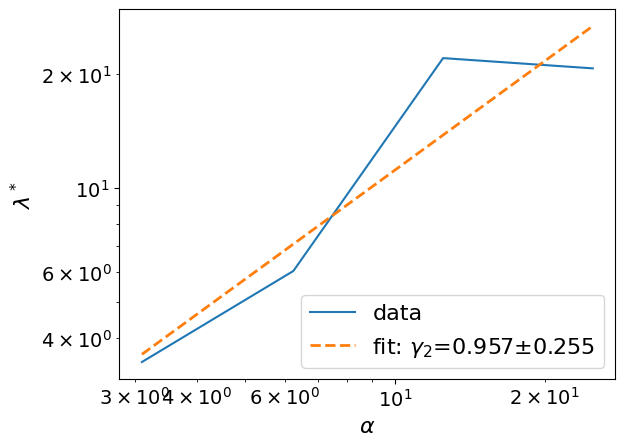}
    \caption{\textbf{Dynamical exponents of unbounded Perceptron are close to the fixed-norm analytical prediction}. (\textit{left}) Curves collapsed by rescaling axes for the minima, using values of $\alpha>25$. (\textit{right}) Fit of the scaling of minima of curves, $\lambda_{opt}(\alpha)$, using only curves for which the minimum have been reached during numerical simulation.}
    \label{fig:fit_perc}
\end{figure}

\section{Training curves in function of time (number of epochs)}
\label{sec:norm_vs_time}
\begin{figure}[H]
    \centering
    \includegraphics[width=0.45\linewidth]{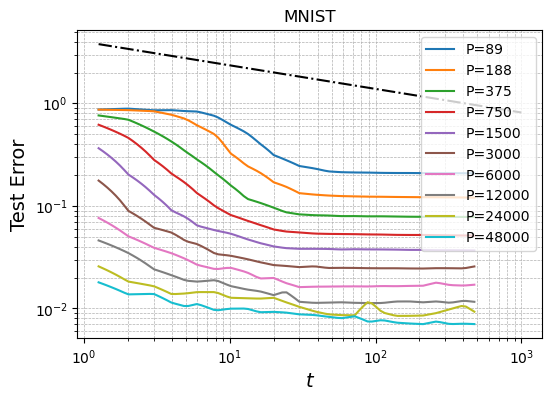}
    \includegraphics[width=0.45\linewidth]{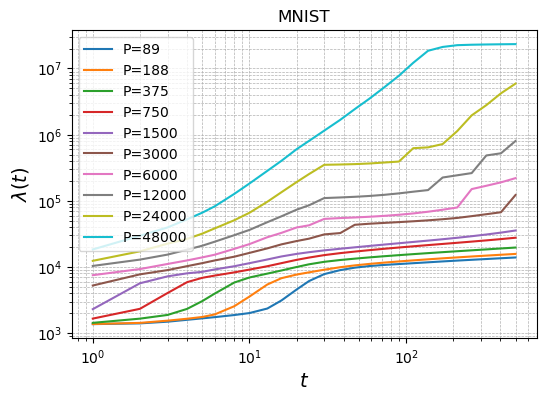}
    \caption{\textbf{The function $\lambda(t)$ is highly non-trivial.} The left panel shows the generalization error of a CNN trained on MNIST as a function of the number of epochs for different dataset sizes $P$. The right panel shows the behavior of the spectral complexity as a function of the number of epochs.}
    \label{fig:norm_vs_time}
\end{figure}

We show in fig. \ref{fig:norm_vs_time} that plotting $\epsilon$ versus time instead of $\lambda$ do not make the curves collapse. In particular $\lambda(t)$ is nonlinear, meaning that the two plots $\epsilon(t)$ and $\epsilon(\lambda)$ are qualitatively different. 

\section{Results of $\epsilon(P)$ power law exponent coefficients and computation of errors}
\label{sec:Hest_emp}
The aim of this section is to explain the procedure used to compute the exponents $\gamma_1, \gamma_2$ of the power laws
\begin{equation*}
    \epsilon = k_1 \lambda^{-\gamma_1} + q_1 ,
\end{equation*}
\begin{equation*}
\lambda_\mathrm{opt} = k_2 P^{\gamma_2} + q_2.    
\end{equation*}
It is possible to combine the two power laws only in the regime of $P$ large enough such that
\begin{equation*}
    \frac{\epsilon}{\epsilon_\mathrm{opt}} = \Phi\left( \frac{\lambda}{\lambda_\mathrm{opt}} \right),
\end{equation*}
with a master curve function $\Phi$ that does not depend on $P$.

The first passage is to decide the minimum $P$ to consider for the procedure. We observed that a value of $P$ slightly bigger or smaller than the chosen one did not change substantially the estimate of $\gamma_1$. In almost all cases we used $P\sim 26000$ as the minimum value. 

\begin{figure}[H]
    \centering
    \includegraphics[width=0.47\linewidth]{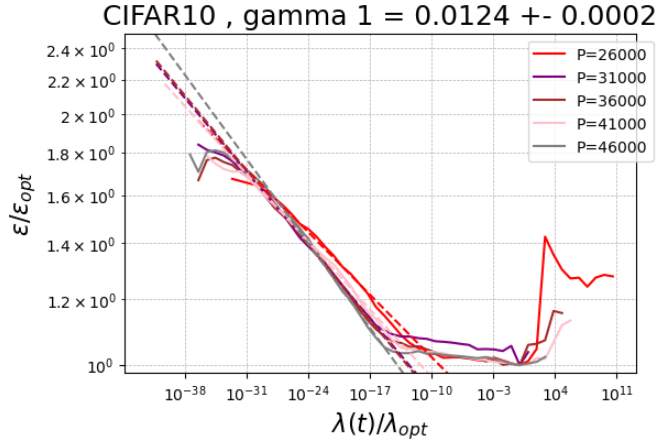}
    \includegraphics[width=0.42\linewidth]{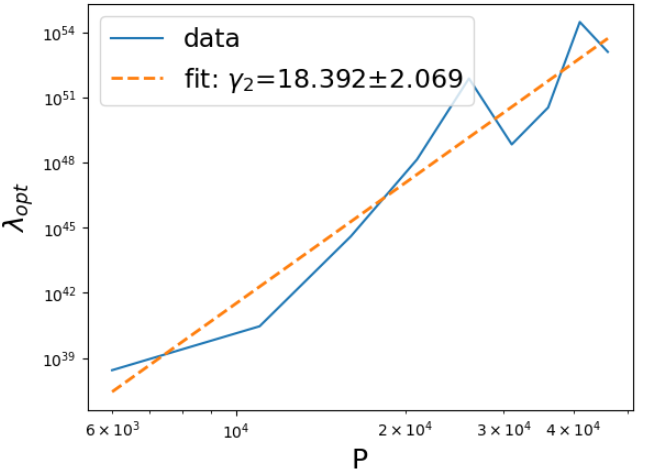}
    \caption{\textbf{The curve collapse helps predict the numerical exponents.} (\textit{left}) Rescaled generalization error curves used to obtain $\gamma_1$ from the fit. The fitted power laws are shown as dashed lines. (\textit{right}) The numerical fit used to estimate $\gamma_2$.}
    \label{fig:passages}
\end{figure}

Then, in the collapsed graph in Fig. \ref{fig:passages} a least-squares fit is performed over the pure power-law region to obtain a prediction of $\gamma_1$ for each value of $P$. The final $\gamma_1$ value is the mean, and the associated error is the error of the mean.

To obtain $\gamma_2$ the minimum of the curves $\lambda^*$ is plotted versus $P$ in Fig. \ref{fig:passages}, and from the fit $\gamma_2$ is obtained with the associated error.

Then $\gamma_\mathrm{pred} = \gamma_1 \gamma_2$ and the error is
\begin{equation*}
    \sigma_\mathrm{pred} = \gamma_\mathrm{pred} \sqrt{\left( \frac{\sigma_1}{\gamma_1}\right)^2+\left( \frac{\sigma_2}{\gamma_2}\right)^2}.
\end{equation*}

\begin{figure}[H]
    \centering
    \includegraphics[width=0.7\linewidth]{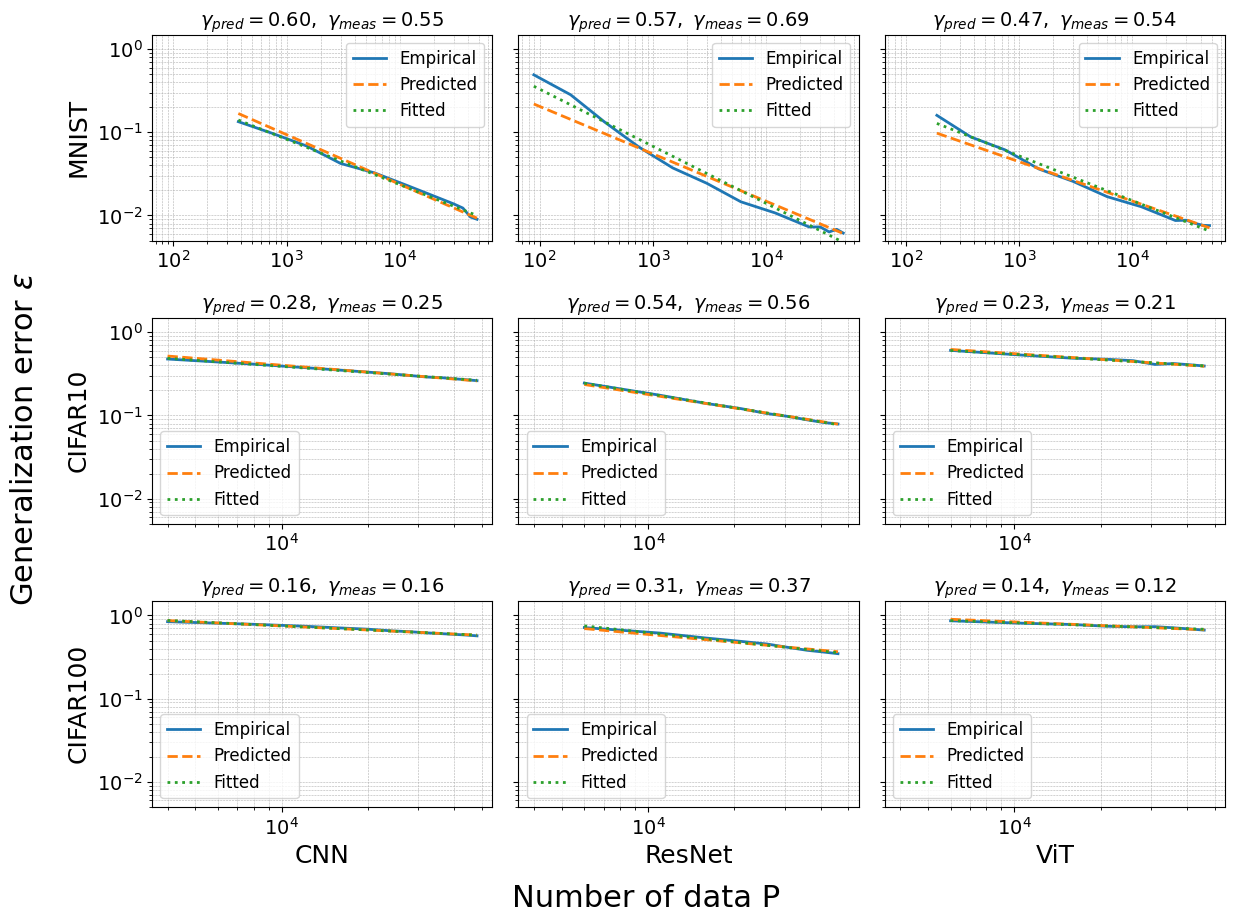}
    \caption{\textbf{The predicted power laws closely match the empirical ones.} We graphically present the numerical results from Table~\ref{tab:error_exponents}. The power laws fitted on the data are compared with the predicted ones. For the predicted power laws, only the exponent is known; the coefficient is chosen to enable visual comparison.}
        \label{fig:Hestness_emp}
\end{figure}

The exponent to compare with is $\gamma_\mathrm{meas}$. For each value of $P$, we considered the minimum of the curve during training, obtaining the empirical curve of $\epsilon(P)$. Then a power-law fit is performed over that curve, obtaining $\gamma_{\textrm{meas}}$ and the $\sigma_\text{meas}$ of the fit. Numerical comparisons are reported in Tab. \ref{tab:error_exponents} and the empirical and predicted power-laws are compared visually in Fig. \ref{fig:Hestness_emp}.
\\ The error assigned to the comparison of exponents is computed as $\sigma = \sqrt{\sigma_\mathrm{pred}^2+\sigma_\mathrm{meas}^2}$. We observe that the magnitude of $\sigma$ is similar across experiments, while exponents change from the maximum of $\gamma_\text{pred}=0.60$ for CNN MNIST to the minimum $\gamma_\text{pred}=0.14$ of ViT CIFAR100. For this reason the relative error is higher the lower is the exponent. Being in possess of more computational power it would be possible to mitigate this effect producing more statistics for models and datasets with lower exponents.

\begin{table}
    \caption{\textbf{Results of the fit for the exponents $\gamma_1$ and $\gamma_2$.} We report the numerical values of the power-law exponents $\gamma_1$ and $\gamma_2$, along with their respective uncertainties, across different datasets and model architectures.}
\centering
\label{tab:error_exponents_2}
\begin{tabular}{llcccc}
\toprule
Model & Dataset & $\gamma_1$ & $\sigma_1$ & $\gamma_2$ & $\sigma_2$ \\
\midrule
CNN & MNIST & 0.59 & 0.06 & 1.01 & 0.11 \\
CNN & CIFAR10 & 0.21 & 0.01 & 1.32 & 0.32 \\
CNN & CIFAR100 & 0.112 & 0.003 & 1.44 & 0.22 \\
ResNet & MNIST & 1.15 & 0.14 & 0.50 & 0.02 \\
ResNet & CIFAR10 & 0.53 & 0.03 & 1.01 & 0.04 \\
ResNet & CIFAR100 & 0.31 & 0.01 & 1.03 & 0.07 \\
ViT & MNIST & 0.139 & 0.005 & 3.41& 0.11 \\
ViT & CIFAR10 & 0.0124 & 0.0002 & 18.4 & 2.1 \\
ViT & CIFAR100 & 0.0068 & 0.0004 & 21 & 6 \\
\bottomrule
\end{tabular}
\end{table}

All intermediate plots as Fig. \ref{fig:passages}, computations and choices of value of $P$ and $\lambda$ to compute power-laws reported in the paper are reported in the supplementary material, as notebooks in the repository of codes with plots and data. We did not report in the paper all details because it would have been necessary to show $\mathcal{O}(100)$ plots to evaluate all cases.

\section{Architectures, datasets, training and resources in details}
\label{sec:spec}

\paragraph{Architectures and hyperparameters}
We used PyTorch Adam optimizer for CNNs and ResNets and AdamW for ViT, in all cases with lerning rate 0.001. We used the standard and most simple possible definitions of the architectures, taken from the original papers. Please refer to the code in the supplementary to the precise definition of each block and width and number of layers. 

\paragraph{Trainings and values of P}
We trained for 500 epochs CNNs and for 1000 epochs ResNets and ViTs. Values of P are
\begin{itemize}
    \item For MNIST in all cases 89, 188, 375, 750, 1500, 3000, 6000, 12000, 24000, 30000, 36000, 42000, 48000.
    \item For CIFAR10 and CIFAR100 with ResNet and ViT in all cases: from 6000 to 46000 every 5000.
    \item For CNN in CIFAR10/100, in the main analysis from 4000 to 48000 every 4000, and in computation of norms and the effect of weight decay from 6000 to 46000 every 5000.
\end{itemize}

\paragraph{Resources to replicate the study}
For perceptron curves the necessary resources are irrelevant. All deep network trainings have been carried on 18 V100 GPUs using 4 CPUs for each, for a period of two months. We set a maximum number of 30 repetitions for each training to get a statistic of learning curves and a month of computation. For smaller models we finished all 30 repetitions while for the slowest one we obtain a total of 5 repetitions.

\section{Effect of weight decay}
\begin{figure}[H]
    \centering
    \includegraphics[width=0.7\linewidth]{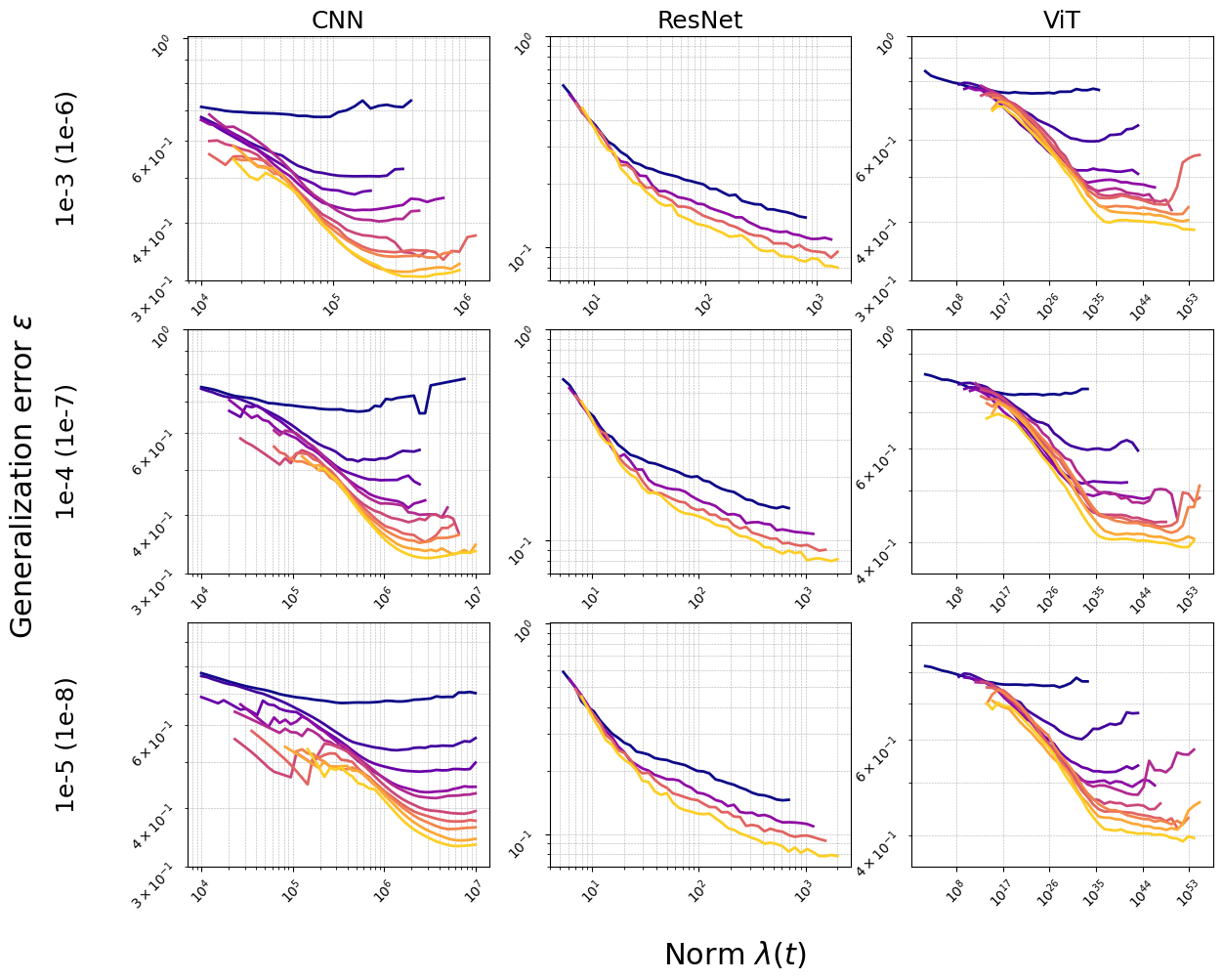}
    \caption{Curves from experiments with weight decay on CIFAR10 dataset. Values of weight decay in parentheses refer to ResNet.}
    \label{fig:eps_vs_lambda_wd}
\end{figure}

\begin{figure}[H]
    \centering
    \includegraphics[width=0.7\linewidth]{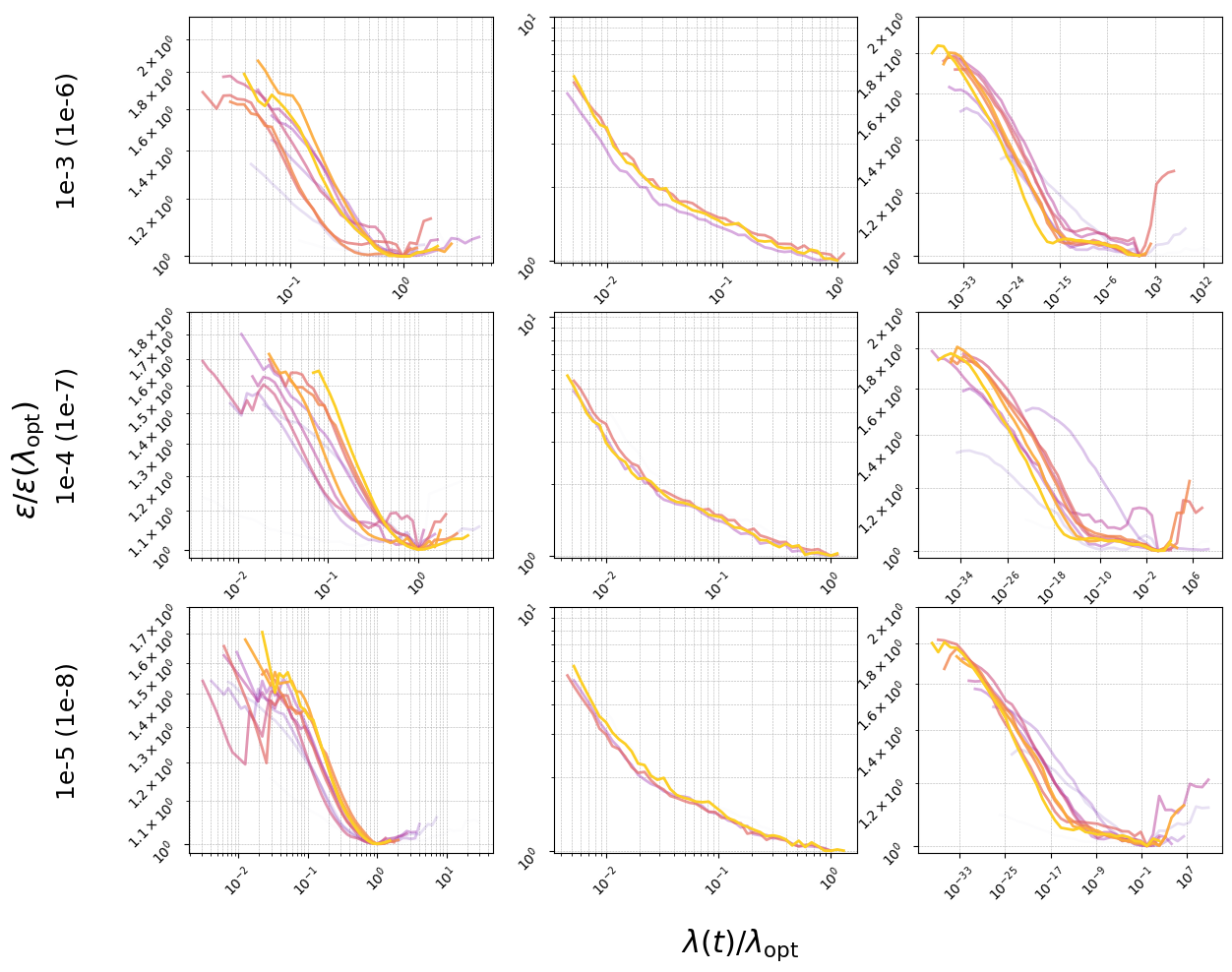}
    \caption{Curves after rescaling collapse onto a master curve also in the presence of a moderate weight-decay. Values of weight decay in parentheses refer to ResNet.}
    \label{fig:phi_wd}
\end{figure}

We reapeted the experiment with the 3 deep architectures analyzed in the paper over CIFAR10 dataset, but with an increasing level of weight decay (WD). In Fig. \ref{fig:eps_vs_lambda_wd} and \ref{fig:phi_wd} we see that in all cases the qualitative picture remain the same, even if the norm of the models doesn't increase monotonically as the case without a weight-decay. 
In Tab. \ref{tab:WD} we observe that the values of $\gamma_1$ and $\gamma_2$ exponents change depending on the amount of weight decay, but their product $\gamma_{pred}$ remains compatible with $\gamma_{meas}$ within the accuracy permitted by the fitting procedure. For ResNet architecture, with a fixed computing budget we found difficult to find the right hyperparameters to obtain overfitting or to saturate the generalization error with the weight decays used for other two architectures, so we reported the result for smaller weight-decays. Due to the increase in training time and corresponding decrease in statistics, the exponents fitted and predicted are affected by a larger error than in other two cases.

\begin{table}[h!]
\centering
\begin{minipage}{0.43\textwidth}
\centering
\resizebox{\textwidth}{!}{%
\begin{tabular}{llccc|}
\toprule
Model & WD & $\gamma_{pred}$ & $\gamma_{meas}$ & $\sigma$ \\
\midrule
CNN & 1e-3 & 0.163 & 0.212 & 0.033 \\
CNN & 1e-4 & 0.136 & 0.193 & 0.050 \\
CNN & 1e-5  & 0.133 & 0.184 & 0.024 \\
ResNet & 1e-6 & 0.269 & 0.525 & 0.090 \\
ResNet & 1e-7 & 0.611 & 0.550 & 0.079 \\
ResNet & 1e-8 & 0.450 & 0.567 & 0.075 \\
ViT & 1e-3 & 0.205 & 0.176 & 0.014 \\
ViT & 1e-4 & 0.198 & 0.174 & 0.023 \\
ViT & 1e-5 & 0.193 & 0.173 & 0.016 \\
\bottomrule
\end{tabular}}
\end{minipage}\hfill
\begin{minipage}{0.57\textwidth}
\centering
\resizebox{\textwidth}{!}{%
\begin{tabular}{llcccc}
\toprule
Model & WD & $\gamma_1$ & $\sigma_1$ & $\gamma_2$ & $\sigma_2$ \\
\midrule
CNN & 1e-3 & 0.2773 & 0.0184 & 0.5883 & 0.0943 \\
CNN & 1e-4 & 0.1880 & 0.0133 & 0.7257 & 0.2551 \\
CNN & 1e-5 & 0.1343 & 0.0177 & 0.9906 & 0.0815 \\
ResNet & 1e-6 & 0.6487 & 0.0247 & 0.4149 & 0.1342 \\
ResNet & 1e-7 & 0.6572 & 0.0298 & 0.9298 & 0.1101 \\
ResNet & 1e-8 & 0.6641 & 0.0272 & 0.6780 & 0.1047 \\
ViT & 1e-3 & 0.0132 & 0.0003 & 15.5590 & 0.7670 \\
ViT & 1e-4 & 0.0121 & 0.0003 & 16.3150 & 1.8161 \\
ViT & 1e-5 & 0.0124 & 0.0003 & 15.5182 & 1.0233 \\
\bottomrule
\end{tabular}}
\end{minipage}
\caption{\textbf{Results on CIFAR10 dataset and increasing levels of weight decay}. (\textit{left}) Predicted and measured exponents. (\textit{right}) $\gamma_1$ and $\gamma_2$ exponents computed by fitting the data.}
\label{tab:WD}
\end{table}
\label{sec:WD}

\begin{figure}[H]
    \centering
    \includegraphics[width=0.7\linewidth]{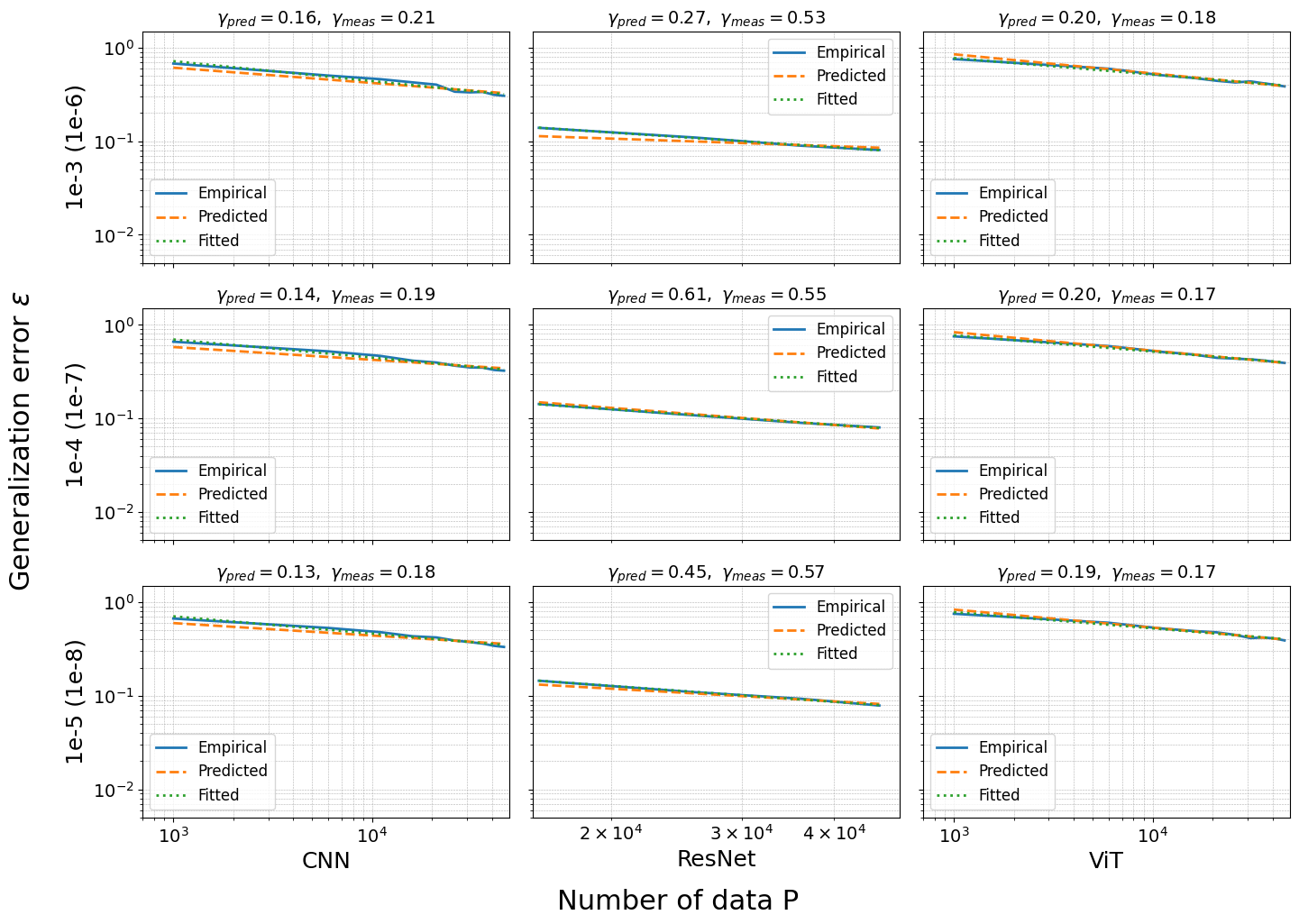}
    \caption{Comparison between predicted scaling laws by combining $\gamma_1$ and $\gamma_2$ and the empirical one measured at end-of-training. Values of weight decay in parentheses refer to ResNet.}
    \label{fig:phi_collapsed_wd}
\end{figure}

\section{Using SGD optimizer instead of Adam on CNNs}
\label{sec:SGD}
\begin{figure}[H]
    \centering
    \includegraphics[width=0.7\linewidth]{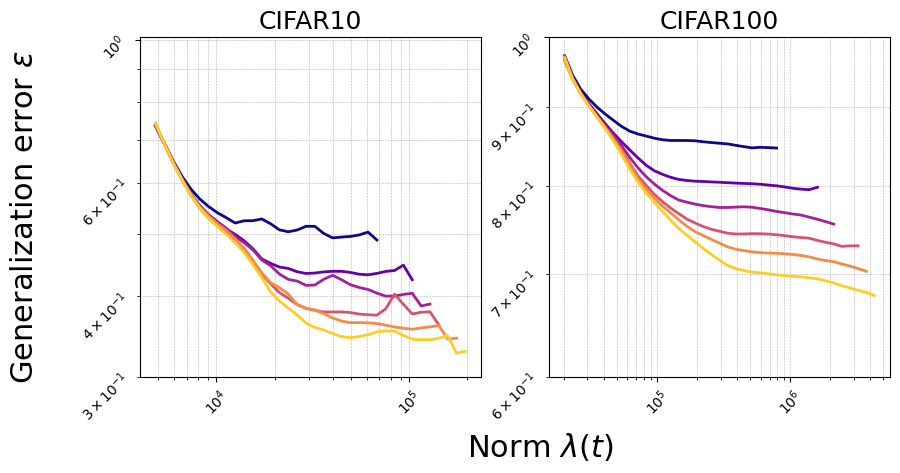}
    \caption{Curves from experiments using SGD optimizer instead of Adam with CNN architecture.}
    \label{fig:eps_vs_lambda_SGD}
\end{figure}

\begin{figure}
    \centering
    \includegraphics[width=0.7\linewidth]{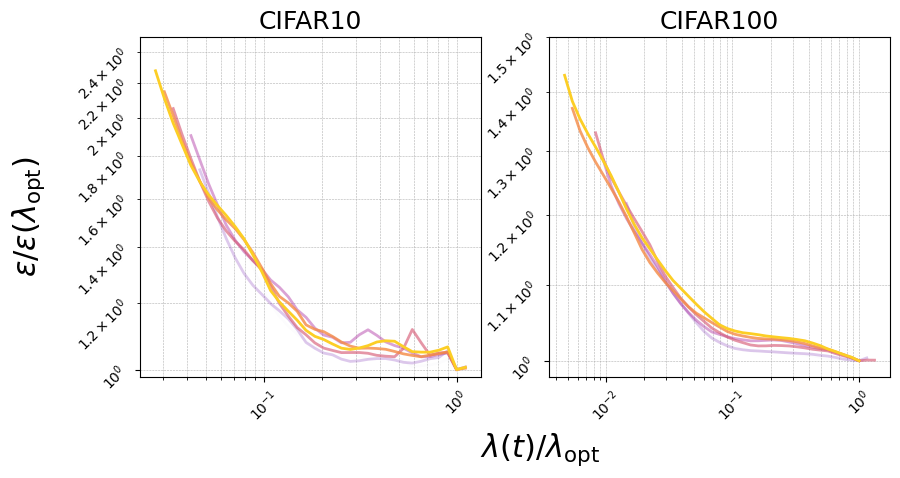}
    \caption{Curves after rescaling collapse onto a master curve also in the case of SGD optimizer instead of Adam.}
    \label{fig:phi_SGD}
\end{figure}

We reapeted the experiment using SGD optimizer instead of Adam, with only CNN architecture over CIFAR10 and CIFAR100 datasets. We did not repeat the experiment over the other two more complex architecture (ResNet, ViT) because Adam and AdamW (respectively used for ResNets and ViTs) are fundamental to make these architectures work appropriately. In Fig. \ref{fig:eps_vs_lambda_SGD} and \ref{fig:phi_SGD} we see that in all cases the qualitative picture remain the same as in the main analysis also for these other norm definitions. At same time exponents predicted are compatible with the ones measured, and compatible as well with the exponents measured in the main analysis using Adam optimizer. This experimental result suggests that the optimizer is not relevant for the end-of-training scaling law exponent $\gamma$, in $\epsilon \sim \epsilon^\gamma$. This instead is not true for the dynamics to reach the optimal value of weights: for example the number of epochs increases dramatically using SGD instead of Adam. This difference in the dynamics is captured from the dynamical exponent. Even though $\gamma_{pred}=\gamma_1 \gamma_2$ is equal with Adam and SGD, we observe that $\gamma_1^{\textrm{SGD}}>\gamma_1^{\textrm{Adam}}$, while $\gamma_2^{\textrm{SGD}}<\gamma_2^{\textrm{Adam}}$.

\begin{table}[H]
\centering
\begin{minipage}{0.445\textwidth}
\centering
\resizebox{\textwidth}{!}{%
\begin{tabular}{llccc|}
\toprule
Model & Norm & $\gamma_{pred}$ & $\gamma_{meas}$ & $\sigma$ \\
\midrule
CNN & CIFAR10 & 0.202 & 0.225 & 0.047 \\
CNN & CIFAR100 & 0.150 & 0.122 & 0.013 \\
\bottomrule
\end{tabular}}
\end{minipage}\hfill
\begin{minipage}{0.555\textwidth}
\centering
\resizebox{\textwidth}{!}{%
\begin{tabular}{llcccc}
\toprule
Model & Norm & $\gamma_1$ & $\sigma_1$ & $\gamma_2$ & $\sigma_2$ \\
\midrule
CNN & CIFAR10 & 0.4735 & 0.0268 & 0.4276 & 0.0962 \\
CNN & CIFAR100 & 0.1469 & 0.0098 & 1.0233 & 0.0170 \\
\bottomrule
\end{tabular}}
\end{minipage}
\caption{\textbf{Results on CIFAR10/100 datasets with CNN using SGD optimizer}. (\textit{left}) Predicted and measured exponents. (\textit{right}) $\gamma_1$ and $\gamma_2$ exponents computed by fitting the data.}
\end{table}

\begin{figure}[H]
    \centering
    \includegraphics[width=0.7\linewidth]{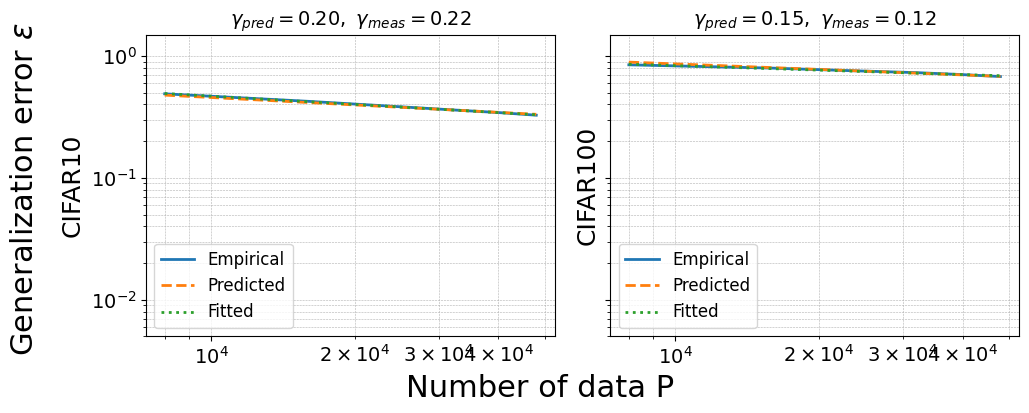}
    \caption{Comparison between predicted scaling laws by combining $\gamma_1$ and $\gamma_2$ and the empirical one measured at end-of-training.}
    \label{fig:placeholder}
\end{figure}

\section{Using other definitions of norm $\lambda$}
\begin{figure}[H]
    \centering
    \includegraphics[width=0.65\linewidth]{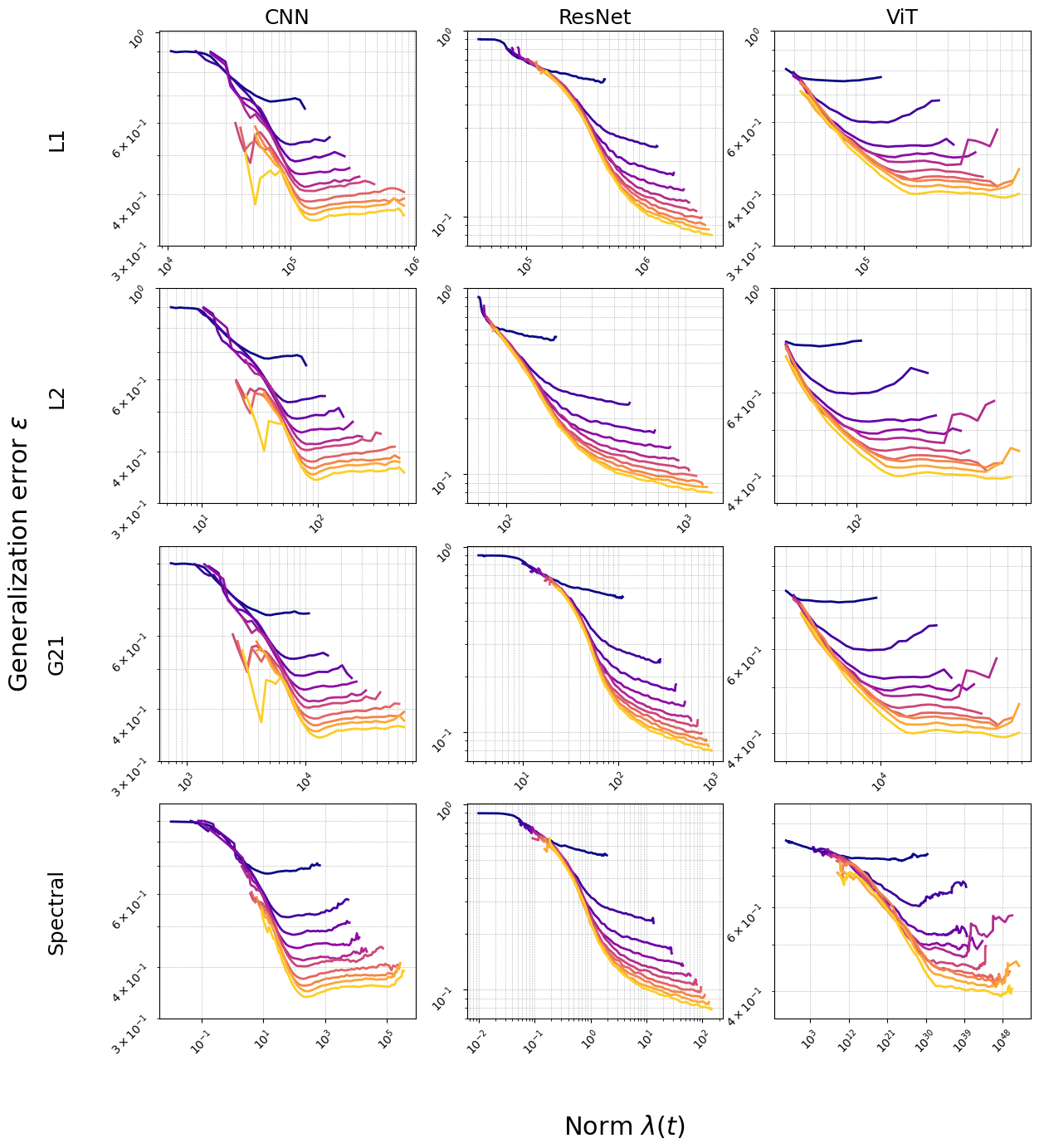}
    \caption{Curves from experiments with different norm definitions on CIFAR10 dataset.}
    \label{fig:eps_vs_lambda_norms}
\end{figure}

\begin{figure}[H]
    \centering
    \includegraphics[width=0.65\linewidth]{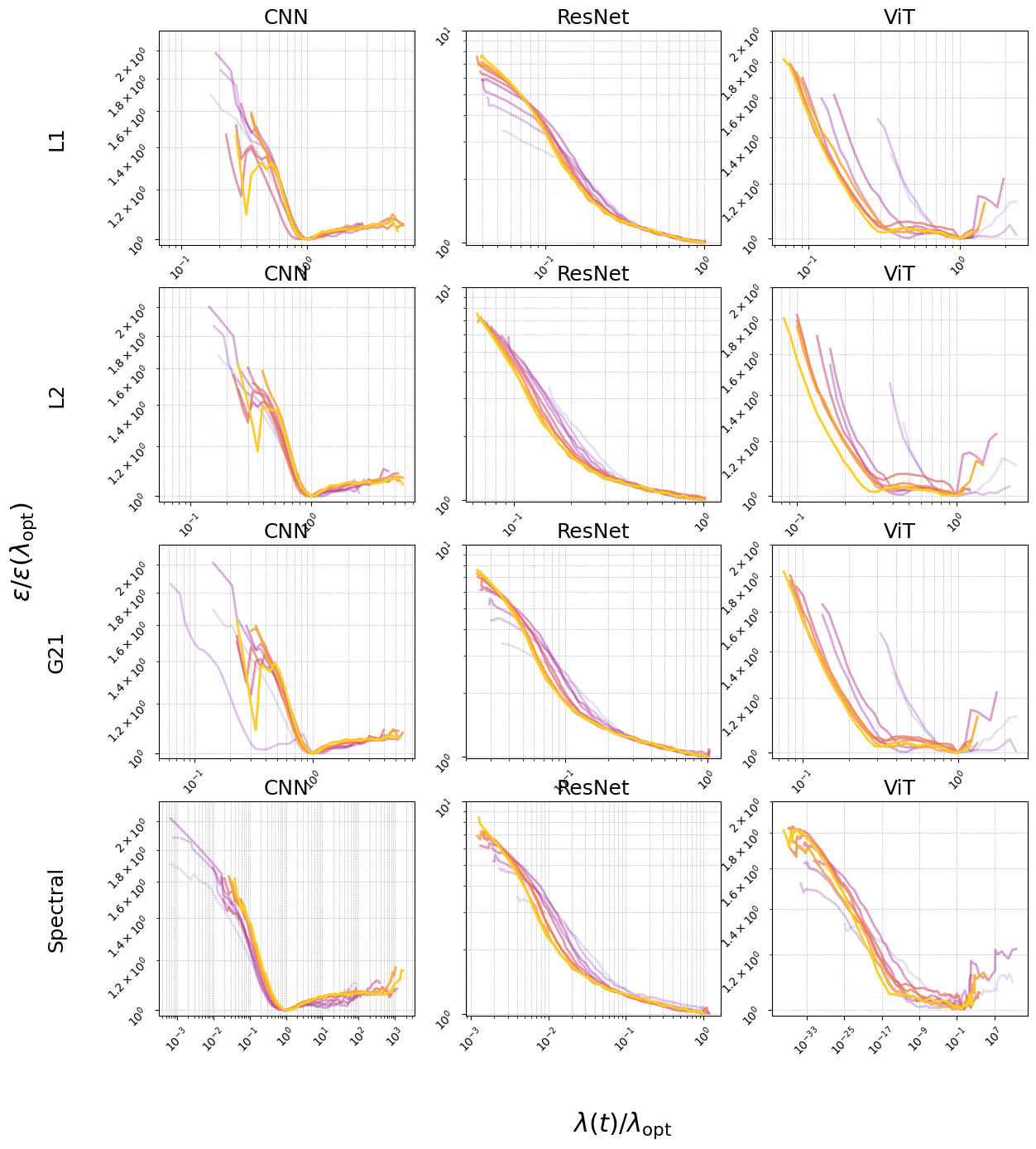}
    \caption{Curves after rescaling collapse onto a master curve also for the other norm considered.}
    \label{fig:phi_norms}
\end{figure}

We reapeted the experiment with the 3 deep architectures analyzed in the paper over CIFAR10 dataset, but measuring other norms instead of the spectral complexity. The norms are:
\begin{enumerate}
\item (L1) Entry-wise $\ell_1$ norm: $ 
  \|A\|_{1} = \sum_{i=1}^{L} \sum_{j,k} |(A_i)_{j,k}|$
\item (L2) Frobenius (entry-wise $\ell_2$) norm: $
  \|A\|_{F} = \left( \sum_{i=1}^{L} \sum_{j,k} (A_i)_{j,k}^2 \right)^{1/2}$
\item (G21) Group $(2,1)$ norm $
  \|A\|_{2,1} = \sum_{i=1}^{L} \sum_{j} \left( \sum_{k} (A_i)_{k,j}^2 \right)^{1/2} $
   i.e.\ the sum over columns of their $\ell_2$ norms.
\item (Spectral) norm product: $
  \prod_{i=1}^{L} \|A_i\|_{\sigma}$, where $\|A_i\|_{\sigma}$ is the largest singular value of $A_i$.
\end{enumerate}
In Fig. \ref{fig:eps_vs_lambda_norms} and \ref{fig:phi_norms} we see that in all cases the qualitative picture remain the same as in the main analysis also for these other norm definitions: plotting learning curves against every tested definition of norm produces the two scaling laws with exponents $\gamma_1$ and $\gamma_2$, and rescaling by minima make the curves to collapse over a master curve. However, the exponents predicted are not compatible with the ones measured. This result suggest that Spectral Complexity norm of Eq. \ref{eq:Bartlett2017} is the correct quantity that generalizes in deep networks the role of L2 norm in the Perceptron analysis. 
\\Even if $\gamma_{meas}\neq\gamma_{pred}=\gamma_1\gamma_2$, we observe a compensation mechanism between $\gamma_1$ and $\gamma_2$ exponents: a bigger $\gamma_1$ implies in almost all cases a smaller $\gamma_2$ with respect to other norms for the same model. 
\begin{table}[h!]
\centering
\begin{minipage}{0.436\textwidth}
\centering
\resizebox{\textwidth}{!}{%
\begin{tabular}{llccc|}
\toprule
Model & Norm & $\gamma_{pred}$ & $\gamma_{meas}$ & $\sigma$ \\
\midrule
CNN & L1 & 0.083 & 0.181 & 0.023 \\
CNN & L2 & 0.118 & 0.181 & 0.028 \\
CNN & G21 & 0.107 & 0.181 & 0.026 \\
CNN & Spectral & 0.081 & 0.181 & 0.042 \\
ResNet & L1 & 0.634 & 0.500 & 0.013 \\
ResNet & L2 & 0.750 & 0.500 & 0.013 \\
ResNet & G21 & 0.680 & 0.500 & 0.018 \\
ResNet & Spectral & 0.641 & 0.500 & 0.011 \\
ViT & L1 & 0.252 & 0.175 & 0.027 \\
ViT & L2 & 0.323 & 0.175 & 0.040 \\
ViT & G21 & 0.262 & 0.175 & 0.028 \\
ViT & Spectral & 0.193 & 0.175 & 0.018 \\
\bottomrule
\end{tabular}}
\end{minipage}\hfill
\begin{minipage}{0.564\textwidth}
\resizebox{\textwidth}{!}{%
\centering
\begin{tabular}{llcccc}
\toprule
Model & Norm & $\gamma_1$ & $\sigma_1$ & $\gamma_2$ & $\sigma_2$ \\
\midrule
CNN & L1 & 0.5687 & 0.0300 & 0.1458 & 0.0358 \\
CNN & L2 & 0.5894 & 0.0157 & 0.2000 & 0.0426 \\
CNN & G21 & 0.5482 & 0.0187 & 0.1958 & 0.0428 \\
CNN & Spectral & 0.1861 & 0.0041 & 0.4339 & 0.2161 \\
ResNet & L1 & 1.1634 & 0.0107 & 0.5447 & 0.0087 \\
ResNet & L2 & 1.4406 & 0.0130 & 0.5208 & 0.0063 \\
ResNet & G21 & 1.1997 & 0.0157 & 0.5669 & 0.0121 \\
ResNet & Spectral & 0.5699 & 0.0058 & 1.1239 & 0.0119 \\
ViT & L1 & 0.4089 & 0.0171 & 0.6170 & 0.0556 \\
ViT & L2 & 0.5491 & 0.0392 & 0.5881 & 0.0571 \\
ViT & G21 & 0.4313 & 0.0173 & 0.6075 & 0.0567 \\
ViT & Spectral & 0.0133 & 0.0001 & 14.4951 & 1.1217 \\
\bottomrule
\end{tabular}}
\end{minipage}
\caption{\textbf{Results on CIFAR10 dataset and different norm definitions}. (\textit{left}) Predicted and measured exponents are not compatible using these norm definitions instead of Spectral Complexity norm. (\textit{right}) $\gamma_1$ and $\gamma_2$ exponents computed by fitting the data.}
\end{table}

\label{sec:other_norms}

\begin{figure}[H]
    \centering
    \includegraphics[width=0.65\linewidth]{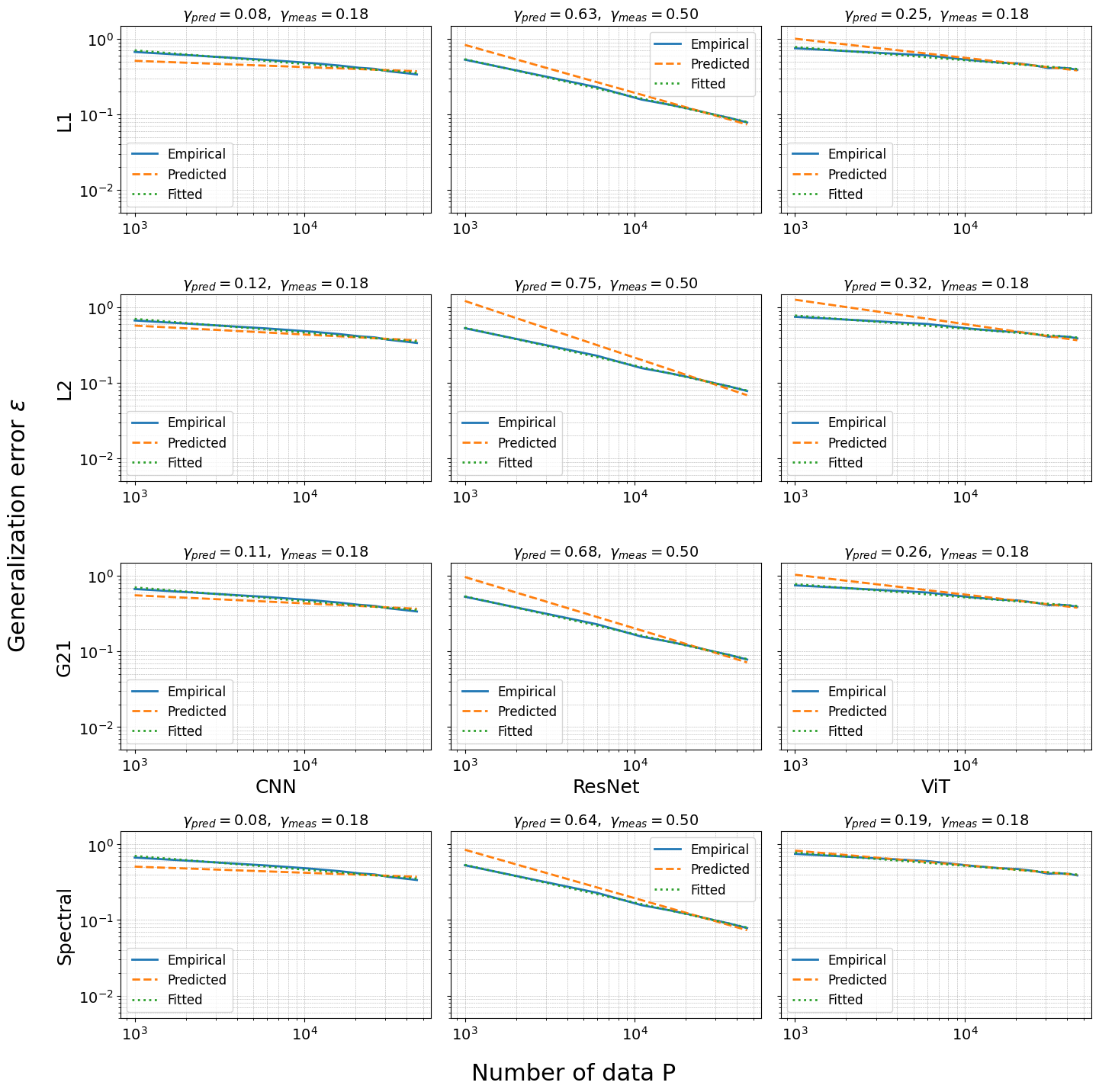}
    \caption{Comparison between predicted scaling laws by combining $\gamma_1$ and $\gamma_2$ and the empirical one measured at end-of-training. Other norms considered predict exponents at end-of-training not always compatible with the empirical ones, even if we can consider them as an approximation of the correct exponent that can be computed using spectral complexity as the norm $\lambda$.}
    \label{fig:fit_final_norms}
\end{figure}

\end{document}